\PassOptionsToPackage{table}{xcolor}
\documentclass{article} 
\usepackage{iclr2027_conference,times}

\usepackage{amsmath,amsfonts,bm}

\def\eqref#1{equation~\ref{#1}}

\def\1{\bm{1}}

\DeclareMathAlphabet{\mathsfit}{\encodingdefault}{\sfdefault}{m}{sl}
\SetMathAlphabet{\mathsfit}{bold}{\encodingdefault}{\sfdefault}{bx}{n}

\usepackage{hyperref}
\usepackage{url}

\usepackage{booktabs}
\usepackage{multirow}
\usepackage{tabularx}
\usepackage{amssymb}   
\usepackage{pifont}    
\usepackage{graphicx}
\usepackage{rotating}
\usepackage{subcaption}
\usepackage{xspace}

\usepackage{booktabs}
\usepackage{multirow}
\usepackage{subcaption}
\usepackage{makecell}
\usepackage{pifont}
\usepackage{graphicx}
\usepackage{tikz}
\usepackage{cleveref}
\usepackage{enumitem}

\newcommand{\mycmark}{\textcolor{green!60!black}{\ding{51}}}
\newcommand{\myxmark}{\textcolor{red!75!black}{\ding{55}}}

\definecolor{headergray}{RGB}{245,247,250}
\definecolor{lightblue}{RGB}{235,244,255}
\definecolor{bestblue}{RGB}{24,84,160}

\title{IVT-Guard: All-in-One Reasoning Model for AI-Generated Content Detection}

\author{\textbf{Hongwei Niu}$^{1,*}$,
\textbf{Yunpeng Luo}$^{2}$,
\textbf{Hanjun Li}$^{2}$,
\textbf{Ziyin Zhou}$^{1}$,
\textbf{Jianghang Lin}$^{1}$, \\
\textbf{Ke Yan}$^{2,\dagger}$,
\textbf{Shouhong Ding}$^{2}$,
\textbf{Shengchuan Zhang}$^{1}$,
\textbf{Liujuan Cao}$^{1,\dagger}$ \\
$^{1}$Key Laboratory of Multimedia Trusted Perception and Efficient Computing, \\
Ministry of Education of China, Xiamen University, \\
Xiamen 361005, P.R. China \\
$^{2}$Tencent YouTu Lab \\
\texttt{niuhw649@gmail.com} \\
\texttt{\{petterluo,kerwinyan\}@tencent.com}
}

\iclrfinalcopy 
\begin{document}

\maketitle

\begin{abstract}
The rapid proliferation of highly realistic AI-Generated Content (AIGC) necessitates robust and interpretable detection mechanisms.
However, existing detectors are predominantly confined to single modalities and provide binary outputs without reasoning.
While Multimodal Large Language Models (MLLMs) present a promising solution,
their development is constrained by the scarcity of multimodal reasoning data and the reasoning--detection optimization dilemma,
where explicit reasoning supervision can compromise detection accuracy.
To this end, we introduce \textbf{IVT-Set}, a comprehensive dataset comprising over 152K diverse image,
video, and text samples equipped with multi-granularity Chain-of-Thought (CoT) reasoning trajectories.
Based on it, we propose \textbf{IVT-Guard},
a pioneering framework for unified and interpretable AIGC detection across image, video, and text modalities.
Furthermore, to overcome the aforementioned optimization dilemma,
we design a novel three-stage training paradigm: Artifact-Aware Pre-training,
Artifact-to-Evidence Supervised Fine-Tuning via artifact-aware injection,
and Evidence-Verdict Consistency Group Relative Policy Optimization.
Extensive experiments demonstrate that IVT-Guard achieves state-of-the-art detection performance
across in-domain, out-of-domain, and cross-dataset settings while delivering faithful reasoning.
Code and data will be released.
\end{abstract}

\section{Introduction}
\label{sec:intro}
\begin{figure}[!t]
  \centering
  \begin{minipage}[t]{0.55\textwidth}
    \vspace{0pt}
    \centering
    \begin{subfigure}[t]{0.48\linewidth}
      \vspace{0pt}
      \includegraphics[width=\linewidth]{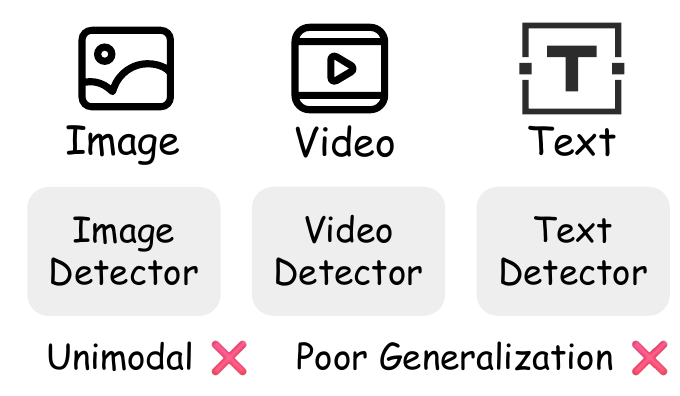}
    \end{subfigure}
    \hfill
    \begin{subfigure}[t]{0.48\linewidth}
      \vspace{0pt}
      \includegraphics[width=\linewidth]{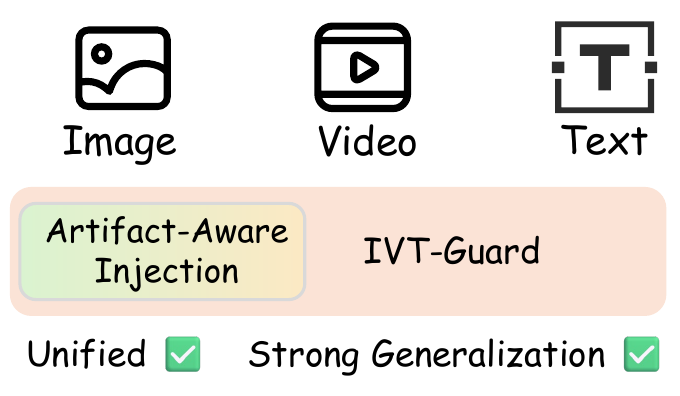}
    \end{subfigure}
    \caption{Comparison of modality-specific methods with our unified detector, IVT-Guard.}
    \label{fig:compare}
  \end{minipage}
  \hfill
\begin{minipage}[t]{0.43\textwidth}
\vspace{0pt}
\centering
\captionof{table}{Comparison of existing datasets.}
\label{tab:datasets_compare}
\scriptsize
\setlength{\tabcolsep}{2.6pt}
\renewcommand{\arraystretch}{1.10}
\resizebox{\linewidth}{!}{%
\begin{tabular}{lcccc}
\toprule
\textbf{Dataset} & \textbf{Mod.} & \textbf{Scale / Gen.} & \textbf{Annotation} & \textbf{Curation} \\
\midrule
FakeClue
& I & 100K / $\geq$5 & CoT & MLLMs \\
ViF-CoT-4K
& V & 4.4K / 7 & CoT & MLLM+Human \\
DetectRL-X
& T & 3.45M / 4 & -- & LLM+Human \\
CommGen15
& I/V & 66.8K / 15 & -- & -- \\
GenBuster-Bench++
& I/V & 4K / 20 & -- & Human \\
LOKI
& I/V/T & 18K / 26 & CoT & Human \\
\midrule
\rowcolor{blue!7}
\textbf{IVT-Set}
& \textbf{I/V/T} & \textbf{152K / 44} & \textbf{M-CoT} & \textbf{MLLMs+Human} \\
\bottomrule
\end{tabular}%
}
\end{minipage}

\end{figure}

\begin{figure}[!t]
  \centering
  \includegraphics[width=\textwidth]{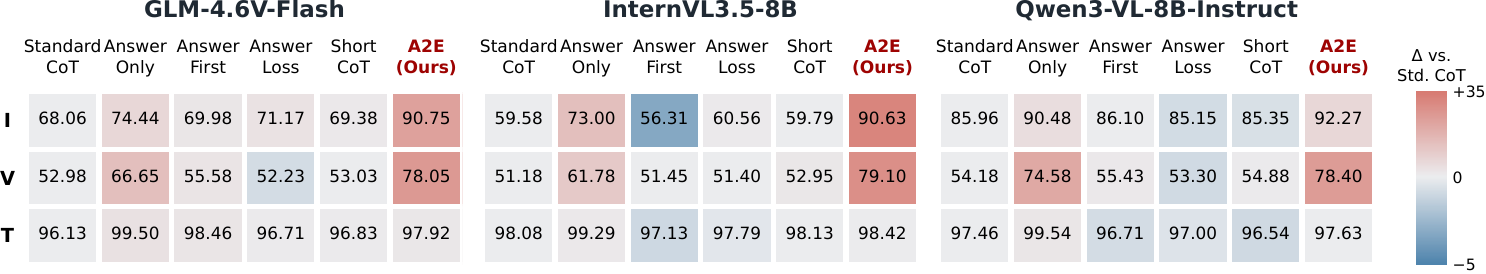}
  \caption{Comparison of OOD accuracy (\%) under different SFT strategies for
  three MLLMs on image (I), video (V), and text (T).
  Colors indicate accuracy changes relative to the standard CoT.}
  \label{fig:sft_ood_heatmap}
\end{figure}

The rapid advancement of generative AI techniques has greatly enriched multimedia content creation.
Modern image generation models~\citep{flux-2-2025,zhao2026qwenimage,krea-2-2026,openai2026gptimage2},
video generation models~\citep{sora2,wu2025hunyuanvideo,wan2025,seedance2026seedance},
multimodal large language models~\citep{Qwen3-VL,google2026gemini31pro,anthropic2026opus5,openai2026gpt56},
and large language models~\citep{qwen3,xu2026deepseek,chen2026minimax,zeng2026glm-5}
can generate highly realistic images, temporally coherent videos, and semantically fluent text,
substantially blurring the boundary between authentic and synthetic content.
In response, substantial progress has been made in AIGC detection.
However, most existing detectors~\citep{yan2025AIDE,2025ReStraV,diveye25,zhou2026pgc,chen2026demamba,liu2026wavedetect}
rely on modality-specific classifiers and typically provide only binary predictions,
limiting both cross-modal generalization and prediction interpretability.
Recent advances in multimodal large language models (MLLMs) have demonstrated strong cross-modal understanding and reasoning capabilities,
providing a promising foundation for unified and interpretable multimodal detection.
Despite this progress, two key challenges remain:
\textbf{1{)} Existing datasets remain siloed}, with none simultaneously providing broad modality coverage,
large-scale data, and multi-granularity Chain-of-Thought (CoT) annotations.
As summarized in~\cref{tab:datasets_compare},
benchmarks such as FakeClue~\citep{wen2025fakevlm},
ViF-CoT-4K~\citep{li2026skyra}, and DetectRL-X~\citep{wu2026detectrl} are limited to individual modalities.
Although CommGen15~\citep{zhou2026pgc}, GenBuster-Bench++~\citep{wen2025busterx++},
and LOKI~\citep{ye2025loki} cover multiple modalities,
they remain limited in scale and generator diversity,
while lacking multi-granularity CoT annotations.
These limitations hinder the development of unified and interpretable AIGC detectors.
\textbf{2{)} Existing MLLMs struggle to maintain both reasoning capabilities and detection performance.}
To analyze this phenomenon, we evaluate six SFT strategies on three MLLMs (GLM-4.6V-Flash, InternVL3.5-8B, and Qwen3-VL-8B-Instruct)
across image, video, and text modalities under out-of-distribution (OOD) settings.
As shown in~\cref{fig:sft_ood_heatmap}, answer-only supervision consistently outperforms standard CoT supervision across all models and modalities.
However, answer-only supervision provides no explicit training signal for structured reasoning.
Several common remedies, such as placing the answer first, reweighting the answer-token loss, or shortening the CoT,
still result in suboptimal detection performance.
%
We attribute this performance trade-off to an optimization dilemma arising from supervised fine-tuning.
Under token-level supervision, lengthy reasoning sequences introduce substantially more language-modeling targets,
diluting the supervision signal of the final verdict and
biasing optimization toward language generation at the expense of modality-specific forensic cues.

%
To address the data limitation, we introduce \textbf{IVT-Set}, a multimodal dataset comprising 152K samples across image, video, and text modalities,
constructed from diverse real-world sources and a wide range of modern generative models.
Through well-crafted prompts and strict filtering mechanisms,
we curate closely matched real-fake pairs to simulate challenging real-world detection scenarios.
Furthermore, we develop an automated annotation pipeline with quality-gated human verification to produce multi-granularity Chain-of-Thought (CoT) trajectories for each sample.
Based on this dataset, we develop \textbf{IVT-Guard}, a unified framework for interpretable AIGC detection.
To resolve the reasoning-detection dilemma in MLLMs and enable reasoning to benefit detection,
we propose a three-stage training paradigm:
(1) \textit{Artifact-Aware Pre-training (AAP):}
We train modality-specific visual encoders with binary classification supervision to capture spatial and temporal artifacts,
yielding artifact-aware forensic features.
(2) \textit{Artifact-to-Evidence Supervised Fine-Tuning (A2E-SFT):}
We introduce an Artifact-Aware Injection mechanism that injects forensic features into the MLLM's visual stream.
CoT-supervised fine-tuning then encourages the model to leverage these features for evidence-grounded reasoning.
(3) \textit{Evidence-Verdict Consistency Group Relative Policy Optimization (EVC-GRPO):}
To further improve detection accuracy and reasoning consistency,
we introduce an evidence-verdict consistency reward that rewards correct and logically consistent reasoning while penalizing contradictions.
As a result, IVT-Guard consistently achieves strong detection performance across in-domain (ID), out-of-domain (OOD), and cross-dataset (CD) settings,
while providing reliable and interpretable evidence for its predictions.

\section{Related Work}
\textbf{AIGC Detection Methods.}
Existing AIGC detection methods are predominantly modality-specific.
For images, methods exploit pretrained vision-language representations~\citep{ojha2023UnivFD,koutlis2024leveraging,liu2024forgery}, pixel-level artifacts~\citep{tan2024NPR},
semantic-forensic features~\citep{yan2025AIDE,cheng2025co}, and local or distributed patch cues~\citep{zhou2026pgc,yang2026all}.
Video detectors model temporal inconsistencies and dynamics~\citep{chen2026demamba,zheng2025d3,zhang2025NSGVD,song2024learning,li2026preserving,corvi2025seeing}.
Text detectors leverage likelihood statistics~\citep{mitchell2023detectgpt,bao2023fast,hans2024Binoculars}, contrastive or diversity-based features~\citep{guo2024detective,diveye25},
and generalization or routing strategies~\citep{fu2025detectanyllm,sun2026minimizing}.

\noindent\textbf{Multimodal Large Language Models for AIGC Detection.}
Recent MLLMs~\citep{Qwen2.5-VL,Qwen3-VL,qwen35blog,wang2025internvl3_5,hong2026glm5v,yu2026minicpmv4_5}
provide a promising foundation for interpretable AIGC detection through multimodal reasoning.
FakeVLM~\citep{wen2025fakevlm} provides artifact-level explanations,
while LEGION~\citep{kang2025legion} and FakeXplainer~\citep{ji2026fakexplain} incorporate artifact localization and visual grounding.
AIGI-Holmes~\citep{zhou2025aigi} further improves generalization and explainability through multi-expert annotation and preference optimization.
BusterX++~\citep{wen2025busterx++} and Skyra~\citep{li2026skyra} extend multimodal explanations and grounded artifact reasoning to video detection,
while VideoVeritas~\citep{tan2026videoveritas} introduces perception-oriented reinforcement learning and
Ivy-Fake~\citep{jiang2026ivy} unifies explainable image-video detection.
However, unified modality coverage and the joint optimization of reasoning and detection remain underexplored.

\section{Dataset}

\subsection{Data Collection}
\textbf{Image.}
We first collect 70K real images from Open Images V7~\citep{kuznetsova2020openimagev7}, AnimeDL~\citep{zhu2025animedl}, CelebA-HQ~\citep{karras2017CelebA-HQ}, CelebA~\citep{liu2015celeba}, and ADE20K~\citep{zhou2017ade20k}.
Constructing challenging real--fake pairs requires visual diversity while preserving semantic and structural similarity.
To this end, we use Q-Insight~\citep{li2025q-insight} for sampling across diverse quality levels,
while Qwen2.5-VL-72B provides category labels and captions for category-balanced sampling and fake-image generation, respectively.
We retain only image pairs with a CLIP~\citep{radford2021clip} semantic similarity above 0.8 and a DINOv2~\citep{oquab2024dinov2} structural similarity above 0.85.
The retained samples are then balanced across resolution, category, and quality score, resulting in approximately 50.5K images.
Finally, we incorporate 3.5K GAN-generated images from WildFake~\citep{hong2025wildfake} to diversify the generation sources.

\noindent\textbf{Video.}
We sample approximately 60K real videos from OpenVid-1M~\citep{nan2025openvid-1m} and ActivityNet Captions~\citep{krishna2017ActivityNet_Captions}, while balancing resolution, frame rate, and frame count.
We filter out low-quality videos using aesthetic scores from the LAION Aesthetic Predictor and
remove temporally inconsistent samples based on the CLIP cosine similarity between adjacent frames.
For fake-video generation, we feed the first frame and caption of each selected real video into diverse open-source and commercial image-to-video models as joint visual and textual prompts.
The generated videos span a wide range of frame rates, durations, and resolutions.
To reduce distribution discrepancies between real and fake videos, we proportionally rescale them for spatial alignment and encode all videos in YUV420p using x265 HEVC.
We further incorporate 5K fake videos from GenBuster-200K~\citep{2025BusterX} and GenVideo~\citep{chen2026demamba}, resulting in approximately 49.4K videos in total.

\noindent\textbf{Text.}
We sample 76K real texts from HC3~\citep{guo-etal-2023-hc3}, Newsroom~\citep{grusky2018newsroom}, SQuAD~\citep{rajpurkar2016squad}, WritingPrompts~\citep{fan2018WritingPrompts}, Arxiv-10~\citep{farhangi2022arxiv10}, and Yelp~\citep{zhang2015yelp}, while balancing word-frequency and domain-label distributions.
The corpus spans question answering, news, encyclopedic content, stories, scientific articles, and reviews.
We first filter out overly short or invalid samples and then perform embedding-based deduplication and text normalization.
For fake-text generation, we employ ten LLMs for question answering, summarization, and rewriting.
Finally, we compute the cosine similarity between real and fake texts using INSTRUCTOR~\citep{su2023INSTRUCTOR} embeddings and
retain pairs with high similarity and closely matched word counts, yielding approximately 48.6K samples.

\subsection{Data Annotation}
\label{sec:annotation}
\begin{figure}[!t]
  \centering
  \includegraphics[width=\textwidth]{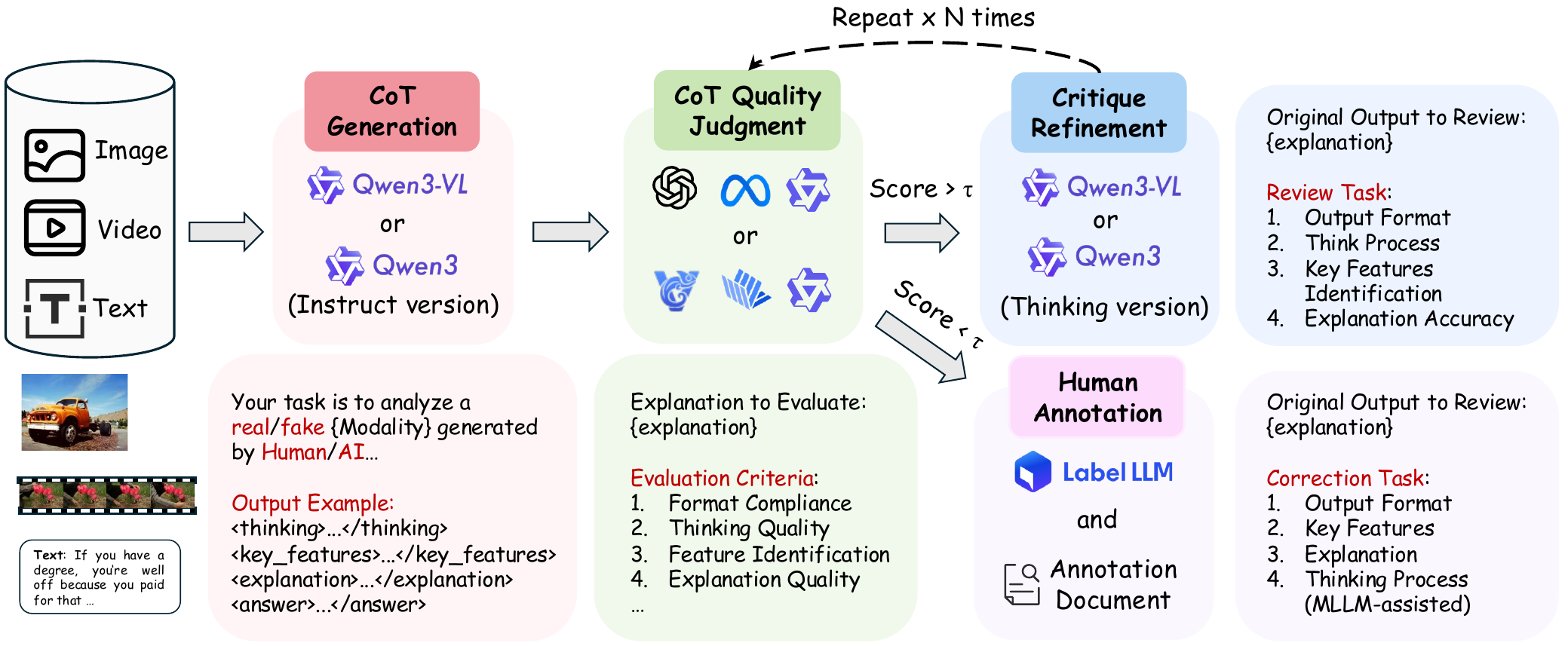}
  \caption{Data annotation pipeline for IVT-Set, consisting of CoT generation,
multi-judge quality judgment, and iterative critique--refinement to produce
high-quality multi-granularity CoT.}
  \label{fig:annotation_pipeline}
\end{figure}

To equip each sample with high-quality CoT annotations at multiple granularities,
we design a three-stage annotation pipeline, as illustrated in~\cref{fig:annotation_pipeline}.
The pipeline consists of the following stages:

\noindent\textbf{CoT Generation}. For each sample, we provide its ground-truth label as a guidance signal
and prompt Qwen3-VL-235B-A22B-Instruct~\citep{Qwen3-VL} to generate an initial CoT annotation.
Each annotation is structured into four components:
\textit{thinking} (step-by-step reasoning), \textit{key features} (salient forensic cues), \textit{explanation} (evidence interpretation), and \textit{answer} (final prediction).
This multi-granularity CoT design establishes a traceable evidence chain from forensic cues to the final verdict, supporting interpretable and evidence-grounded reasoning.

\noindent\textbf{CoT Quality Judgment}. We employ multiple MLLMs or LLMs as independent judges to evaluate annotations across several quality dimensions and compute the average score.
Annotations below the review threshold are assigned to human annotators for revision following standardized guidelines,
while the others proceed to the Critique--Refinement Loop.

\noindent\textbf{Critique--Refinement Loop}. In this stage, we use Qwen3-VL-235B-A22B-Thinking to critique and refine annotations following the quality rubric from the previous stage,
and re-evaluate them with the same MLLM/LLM judges.
The critique--refinement--evaluation cycle continues until the average score meets the acceptance threshold or three refinement rounds are completed.

\section{Method}
\label{sec:method_first}
\begin{figure}[!t]
  \centering
  \includegraphics[width=\textwidth]{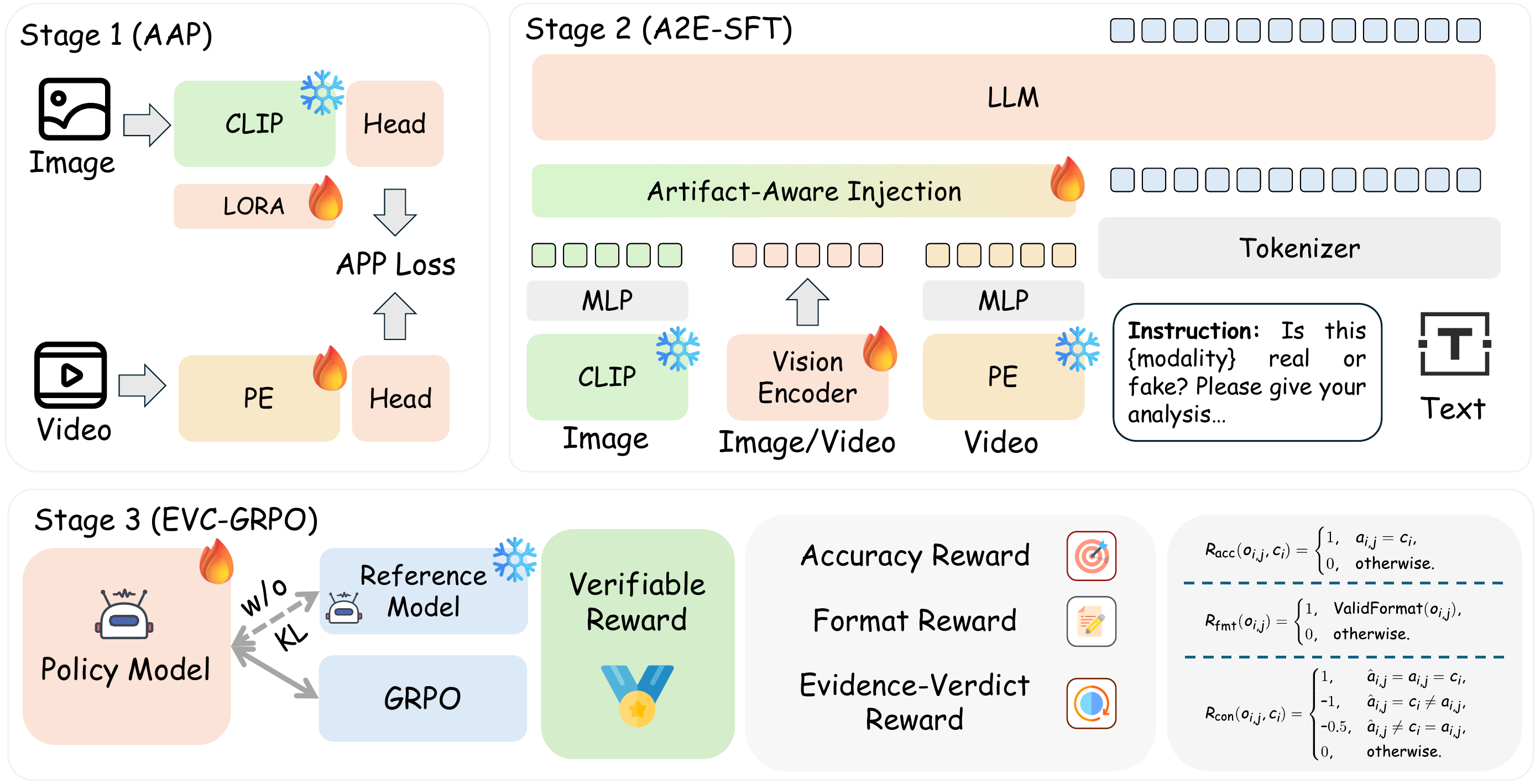}
    \caption{Overview of the three-stage training pipeline for IVT-Guard.
  Stage 1: AAP enhances sensitivity to modality-specific generation artifacts;
  Stage 2: A2E-SFT integrates artifact-aware representations into the MLLM for evidence-grounded reasoning;
  and Stage 3: EVC-GRPO jointly optimizes detection accuracy and evidence--verdict consistency.}
  \label{fig:train_pipeline}
\end{figure}

To jointly optimize detection accuracy and reasoning capabilities, we propose a three-stage training paradigm for IVT-Guard,
consisting of Artifact-Aware Pre-training (AAP), Artifact-to-Evidence Supervised Fine-Tuning (A2E-SFT),
and Evidence-Verdict Consistency Group Relative Policy Optimization (EVC-GRPO), as illustrated in~\cref{fig:train_pipeline}.

\subsection{Artifact-Aware Pre-training}
The visual encoders of MLLMs are primarily optimized for high-level semantic understanding
and underrepresent subtle generation artifacts.
We therefore introduce Artifact-Aware Pre-training with two modality-specific perceptual amplifiers for spatial and temporal artifact perception.

\noindent\textbf{Spatial Perceptual Amplifier.}
For an image input $x_i$, we employ the CLIP~\citep{radford2021clip} image encoder and
adapt its pretrained parameters $\boldsymbol{\theta}_0$ with LoRA modules $\Delta\boldsymbol{\theta}$.
The encoder produces a sequence of visual tokens $\mathbf{F}_i$,
from which we extract the \texttt{[CLS]} token as the spatially aggregated artifact feature $\mathbf{f}_i$:
\begin{equation}
    \mathbf{F}_i
    =
    \mathcal{E}_{\mathrm{CLIP}}
    \!\left(
        x_i;\,
        \boldsymbol{\theta}_0 + \Delta\boldsymbol{\theta}
    \right)
    \in
    \mathbb{R}^{(N_c+1)\times d_c},
    \qquad
    \mathbf{f}_i
    =
    \left[\mathbf{F}_i\right]_{0,:}.
    \label{eq:image_feature}
\end{equation}
The Spatial Perceptual Amplifier captures both low-level texture irregularities and high-level semantic inconsistencies in images.

\noindent\textbf{Temporal Perceptual Amplifier.}
Given a video input $v_i$ consisting of $T$ frames $\{v_{i,t}\}_{t=1}^{T}$,
we employ the token encoder of the Perception Encoder (PE)~\citep{bolya2026perception} to extract frame-level visual tokens.
The resulting tokens $\mathbf{H}_{i,t}$ are aggregated by the PE attention pooling module into a frame-level artifact representation for each frame.
We then perform mean pooling over these representations to obtain the temporally aggregated artifact feature $\mathbf{h}_i$:
\begin{equation}
    \mathbf{H}_{i,t}
    =
    \mathcal{T}_{\mathrm{PE}}^{\mathrm{tok}}
    \!\left(
        v_{i,t}
    \right)
    \in
    \mathbb{R}^{(N_p+1)\times d_p},
    \qquad
    \mathbf{h}_i
    =
    \frac{1}{T}
    \sum_{t=1}^{T}
    \operatorname{AttnPool}_{\mathrm{PE}}
    \!\left(
        \mathbf{H}_{i,t}
    \right).
    \label{eq:video_feature}
\end{equation}
The Temporal Perceptual Amplifier captures frame-level visual irregularities and aggregates generation artifacts across video frames.

We separately introduce binary authenticity supervision for the two modality-specific perceptual amplifiers.
Specifically, the spatially aggregated artifact feature $\mathbf{f}_i$ and
temporally aggregated artifact feature $\mathbf{h}_i$ are fed into
their respective classification heads $g_f$ and $g_h$ to predict authenticity.
Both perceptual amplifiers are optimized using the same binary cross-entropy objective:
\begin{equation}
    \mathcal{L}_{\mathrm{AAP}}
    =
    -\frac{1}{B}
    \sum_{i=1}^{B}
    \left[
        y_i\log \sigma\!\left(g(\mathbf u_i)\right)
        +
        (1-y_i)\log\!\left(
            1-\sigma\!\left(g(\mathbf u_i)\right)
        \right)
    \right],
    \;
    (\mathbf u_i,g)\in
    \{(\mathbf f_i,g_f),(\mathbf h_i,g_h)\},
    \label{eq:classification_loss}
\end{equation}
where $\sigma(\cdot)$ denotes the sigmoid function and $y_i\in\{0,1\}$ denotes the ground-truth label.
This supervision encourages both perceptual amplifiers to learn representations sensitive to authenticity-related generation artifacts.

\subsection{Artifact-to-Evidence Supervised Fine-Tuning}
\label{sec:method_second}
To integrate the learned artifact representations into the MLLM's visual stream,
we introduce an \emph{Artifact-Aware Injection} mechanism.
It fuses artifact representations with the native visual tokens at complementary granularities,
supporting evidence-grounded reasoning over modality-specific generation artifacts.

\noindent\textbf{Image Modality: Global-Local Injection.}
For an image $x_i$, the Spatial Perceptual Amplifier produces the spatially aggregated artifact feature $\mathbf{f}_i$
and the local artifact features $\mathbf{F}_i[1:,:]$.
We inject these artifact features into the native MLLM visual tokens $V_i^{\mathrm{mllm}}\in\mathbb R^{N_i\times d}$ through two complementary pathways.
Specifically, the projected global feature is broadcast to all visual tokens to provide holistic image-level artifact context,
while the projected local features are selectively incorporated through cross-attention to capture fine-grained local irregularities:
\begin{equation}
    \widetilde{\mathbf V}_{i}^{\mathrm{img}}
    =
    \mathbf V_{i}^{\mathrm{mllm}}
    +
    \phi_{\mathrm{g}}^{\mathrm{img}}(\mathbf f_i)
    +
    \operatorname{CrossAttn}
    \left(
        \mathbf V_{i}^{\mathrm{mllm}},
        \phi_{\mathrm{l}}^{\mathrm{img}}
        \left(
            \mathbf F_i[1:,:]
        \right)
    \right),
    \label{eq:image_injection}
\end{equation}
where $\phi_{\mathrm{g}}^{\mathrm{img}}$ and $\phi_{\mathrm{l}}^{\mathrm{img}}$ denote the global and local MLP projectors, respectively.
This global-local injection combines holistic image-level cues with fine-grained local irregularities,
providing complementary artifact evidence for subsequent reasoning.

\noindent\textbf{Video Modality: Temporal-Spatial Injection.}
For a video $v_i$, the Temporal Perceptual Amplifier yields the
temporally aggregated artifact feature $\mathbf{h}_i$ together with
frame-specific local artifact features $\mathbf{H}_{i,t}[1:,:]$.
The former captures video-level artifact information aggregated across frames,
whereas the latter retains spatial irregularities within individual frames.
For the $t$-th frame, we inject both into the native MLLM visual tokens $\mathbf V_{i,t}^{\mathrm{mllm}}$:
\begin{equation}
    \widetilde{\mathbf V}_{i,t}^{\mathrm{vid}}
    =
    \mathbf V_{i,t}^{\mathrm{mllm}}
    +
    \phi_{\mathrm{t}}^{\mathrm{vid}}(\mathbf h_i)
    +
    \operatorname{CrossAttn}
    \left(
        \mathbf V_{i,t}^{\mathrm{mllm}},
        \phi_{\mathrm{s}}^{\mathrm{vid}}
        \left(
            \mathbf H_{i,t}[1:,:]
        \right)
    \right),
    \label{eq:video_injection}
\end{equation}
where $\phi_{\mathrm{t}}^{\mathrm{vid}}$ and $\phi_{\mathrm{s}}^{\mathrm{vid}}$ denote the temporal and spatial MLP projectors, respectively.
The temporal feature is shared across frames to provide consistent video-level artifact context,
while cross-attention incorporates frame-specific spatial artifact cues.
The enhanced frame-level visual tokens are concatenated to
form the final video representation:
\begin{equation}
    \widetilde{\mathbf V}_{i}^{\mathrm{vid}}
    =
    \operatorname{Concat}
    \left(
        \widetilde{\mathbf V}_{i,1}^{\mathrm{vid}},
        \ldots,
        \widetilde{\mathbf V}_{i,T}^{\mathrm{vid}}
    \right).
    \label{eq:video_token_concat}
\end{equation}

For text inputs, we use the MLLM's native textual representations.
Finally, we jointly fine-tune the MLLM across image, video, and text modalities with CoT supervision.
Given a batch of target CoT response sequences
$\{Z_i\}_{i=1}^{B}$, where
$Z_i=\{z_{i,1},\ldots,z_{i,L_i}\}$ and $L_i$ denotes the length of the $i$-th response sequence,
we optimize the A2E-SFT objective using autoregressive language modeling:
\begin{equation}
    \mathcal{L}_{\text{A2E-SFT}}
    =
    -\frac{1}{\sum_{i=1}^{B} L_i}
    \sum_{i=1}^{B}
    \sum_{\ell=1}^{L_i}
    \log P_{\theta}\!\left(
        z_{i,\ell}
        \mid
        z_{i,<\ell},
        x_i^{m_i}
    \right),
    \qquad
    m_i\in\{\mathrm{img},\mathrm{vid},\mathrm{txt}\},
    \label{eq:sft_objective}
\end{equation}
where $x_i^{m_i}$ denotes the modality-specific input of sample $i$.
Through supervised fine-tuning,
the MLLM learns to translate artifact-aware representations into
forensic evidence and perform evidence-grounded reasoning.
\subsection{Evidence-Verdict Consistency Group Relative Policy Optimization}
\label{sec:method_third}
In this stage, to reinforce the model's reasoning behavior through outcome-based rewards while improving detection accuracy,
we further optimize the A2E-SFT-initialized policy $\pi_{\theta}$
using Group Relative Policy Optimization (GRPO)~\citep{shao2024deepseekmath,guo2025deepseek}.
For each input-question pair $(x_i,Q_i)$, the current policy samples a group of $G$ responses
$\mathcal{O}_i=\{o_{i,j}\}_{j=1}^{G}$,
where $i$ indexes the input and $j$ indexes the response within the group.
Each response contains a reasoning trace and an explicit verdict
$a_{i,j}\in\{\mathrm{real},\mathrm{fake}\}$.
To jointly encourage accurate predictions, valid response structures, and consistency between the generated reasoning and final verdict,
we design a composite reward consisting of an accuracy reward, a format reward, and an evidence-verdict consistency reward.
Specifically, we first define the accuracy reward and the format reward as follows:
\begin{equation}
    R_{\mathrm{acc}}(o_{i,j},c_i)
    =
    \begin{cases}
        1, & a_{i,j}=c_i,\\
        0, & \text{otherwise},
    \end{cases}
    \qquad
    R_{\mathrm{fmt}}(o_{i,j})
    =
    \begin{cases}
        1, & \operatorname{ValidFormat}(o_{i,j}),\\
        0, & \text{otherwise}.
    \end{cases}
    \label{eq:accuracy_format_reward}
\end{equation}
where $\operatorname{ValidFormat}(o_{i,j})$ indicates whether the response follows the predefined output format,
requiring the tags \texttt{<thinking>}, \texttt{<key\_features>}, \texttt{<explanation>},
and \texttt{<answer>} to appear in the prescribed order.

\noindent\textbf{Evidence-Verdict Consistency Reward.}
To assess whether the generated reasoning supports the final verdict,
we remove the explicit answer and provide the preceding reasoning content,
together with the verification prompt $p_{\mathrm{v}}$, to
Qwen3-4B-Instruct-2507~\citep{qwen3}, which serves as the external LLM judge:
\begin{equation}
    \hat a_{i,j}
    =
    \operatorname{Judge}
    \!\left(
        p_{\mathrm{v}},
        \operatorname{Reason}(o_{i,j})
    \right),
    \qquad
    \hat a_{i,j}
    \in
    \{\mathrm{real},\mathrm{fake},\mathrm{uncertain}\}.
    \label{eq:judge_verdict}
\end{equation}
Using the ground-truth label $c_i$, we define the consistency reward as
\begin{equation}
    R_{\mathrm{con}}(o_{i,j},c_i)
    =
    \begin{cases}
        1,
        & \hat a_{i,j}=a_{i,j}=c_i, \\[2pt]
        -1,
        & \hat a_{i,j}=c_i\neq a_{i,j}, \\[2pt]
        -0.5,
        & \hat a_{i,j}\neq c_i=a_{i,j},
          \ \hat a_{i,j}\in\{\mathrm{real},\mathrm{fake}\}, \\[2pt]
        0,
        & \text{otherwise}.
    \end{cases}
    \label{eq:consistency_reward}
\end{equation}
The consistency reward is positive only when the judge verdict, the model's final verdict,
and the ground-truth label coincide.
A larger penalty is assigned when the reasoning supports the ground-truth verdict but the final verdict is incorrect,
whereas a smaller penalty is applied when the final verdict is correct but inconsistent with the reasoning.
Uncertain judge outputs receive zero consistency reward.
The final reward is defined as
\begin{equation}
    R_{i,j}
    =
    0.9\,R_{\mathrm{acc}}(o_{i,j},c_i)
    +
    0.1\,R_{\mathrm{fmt}}(o_{i,j})
    +
    0.2\,R_{\mathrm{con}}(o_{i,j},c_i).
    \label{eq:evc_reward}
\end{equation}
The policy is optimized using the following EVC-GRPO objective:
\begin{equation}
    \mathcal{L}_{\text{EVC-GRPO}}
    =
    -\mathbb{E}\!\left[
        \frac{1}{G}\sum_{j=1}^{G}
        \frac{1}{|o_{i,j}|}
        \sum_{\ell=1}^{|o_{i,j}|}
        \min\!\left(
            \rho_{i,j,\ell}(\theta)\widehat{A}_{i,j},
            \operatorname{clip}\!\left(
                \rho_{i,j,\ell}(\theta),1-\epsilon,1+\epsilon
            \right)\widehat{A}_{i,j}
        \right)
    \right],
    \label{eq:grpo_objective}
\end{equation}
where $\rho_{i,j,\ell}(\theta)$ denotes the importance ratio for the $j$-th
response of the $i$-th input between the updated and rollout policies, and
$\widehat{A}_{i,j}
=
\dfrac{
R_{i,j}
-
\operatorname{mean}\!\left(\{R_{i,k}\}_{k=1}^{G}\right)
}{
\operatorname{std}\!\left(\{R_{i,k}\}_{k=1}^{G}\right)
}$
denotes its group-relative advantage. We set
$\epsilon=0.2$ and omit the KL-divergence regularization term.

\section{Experiments}
\label{sec:experiments}

\subsection{Experimental Setup}
\noindent\textbf{Implementation details.}
We train IVT-Guard based on Qwen3-VL-8B-Instruct~\citep{Qwen3-VL}.
During training, we uniformly sample eight frames from each video.
In AAP, we adapt CLIP ViT-L/14@336px with rank-8
LoRA and fine-tune the last six blocks of PE-Core-L14-336.
In A2E-SFT, we freeze the CLIP and PE encoders and fine-tune the MLLM with LoRA, setting the LoRA rank to 128.
We use a learning rate of $1\times10^{-5}$ and a global batch size of 128.
For EVC-GRPO, we use AdamW~\citep{loshchilov2019adamw} with a learning rate of $1\times10^{-6}$,
a global batch size of 128, and 8 rollouts per prompt.
More training and inference details are provided in Appendix~\ref{apd:appendix_implementation_details}.

\noindent\textbf{Evaluation Protocols.}
For modality-specific detectors, we retrain each method on IVT-Set following its original training configuration.
For generic off-the-shelf MLLMs and IVT-Guard, we use the same prompt.
For specialized MLLM-based detectors, we follow their original prompting protocols.
We evaluate all methods under in-domain (ID), cross-dataset (CD),
and out-of-domain (OOD) settings, reporting accuracy (\%).
More details are provided in Appendix~\ref{apd:evaluation_data}.

\begin{table}[!t]
\centering
\caption{
Accuracy comparison (\%) on the image modality of IVT-Set and four cross-dataset benchmarks.
\emph{Avg.} denotes the mean of ID accuracy, average CD accuracy, and average OOD accuracy.
Best and second-best results are shown in \textbf{bold} and \underline{underlined}, respectively.
}
\label{tab:image_results}

\resizebox{\textwidth}{!}{%
\begin{tabular}{
l
c
cccc
ccccccc
c
}
\toprule
\multirow[c]{2}{*}{\textbf{Method}}
& \multirow[c]{2}{*}{\textbf{ID}}
& \multicolumn{4}{c}{\textbf{CD}}
& \multicolumn{7}{c}{\textbf{OOD}}
& \multirow[c]{2}{*}{\textbf{\emph{Avg.}}} \\
\cmidrule(lr){3-6}
\cmidrule(lr){7-13}
&
&
\textbf{Chameleon}
& \textbf{LOKI}
& \textbf{GenImage}
& \textbf{GenBuster++}
& \textbf{Real}
& \textbf{FLUX.1 Krea}
& \textbf{Qwen-Image}
& \textbf{Midjourney v6}
& \textbf{Infinity}
& \textbf{Janus-Pro-7B}
& \textbf{LlamaGen}
& \\
\midrule

\rowcolor{gray!12}
\multicolumn{14}{l}{
\textbf{\textit{Image-specific Detectors}}
} \\

UnivFD~\citep{ojha2023UnivFD}
& 77.75
& 52.42
& 67.70
& 72.97
& 54.70
& 82.75
& \underline{79.88}
& 68.00
& 73.38
& 73.62
& 74.12
& 87.38
& 72.24 \\

DIRE~\citep{wang2023DIRE}
& 89.00
& 70.92
& 71.23
& 75.95
& \textbf{72.35}
& 89.75
& 72.88
& 55.75
& 74.75
& 81.50
& 71.00
& 87.25
& 79.25 \\

NPR~\citep{tan2024NPR}
& 84.62
& 60.15
& 66.07
& 75.70
& 52.90
& 91.25
& 61.12
& 54.62
& 70.12
& 81.00
& 62.62
& 65.25
& 72.58 \\

AIDE~\citep{yan2025AIDE}
& 88.62
& 61.81
& 77.74
& 87.76
& 60.15
& 89.25
& 79.75
& 74.12
& 74.25
& 84.12
& 80.62
& 90.00
& 80.74 \\

DGS-Net~\citep{yan2026dgs}
& 84.75
& 54.64
& 75.14
& 84.73
& 49.35
& 76.75
& 78.38
& \underline{88.00}
& 72.50
& 86.25
& 87.88
& 87.88
& 77.75 \\

PGC~\citep{zhou2026pgc}
& \underline{93.12}
& 67.60
& 80.78
& 92.54
& 54.70
& 90.25
& 61.88
& 83.12
& 66.00
& 87.12
& 92.62
& \underline{94.75}
& \underline{83.09} \\

\midrule

\rowcolor{gray!12}
\multicolumn{14}{l}{
\textbf{\textit{Generic MLLMs}}
} \\

Qwen2.5-VL-7B-Instruct
& 54.62
& 63.56
& 50.60
& 53.57
& 51.02
& 95.50
& 52.25
& 50.62
& 51.62
& 52.00
& 52.38
& 65.75
& 56.44 \\

Qwen3-VL-8B-Instruct
& 65.25
& 78.33
& 67.92
& 70.95
& \underline{66.85}
& 89.50
& 55.50
& 53.50
& 55.25
& 72.00
& 66.88
& 91.12
& 68.46 \\

InternVL3.5-8B
& 61.00
& 62.93
& 55.70
& 63.65
& 52.05
& 92.75
& 52.12
& 54.00
& 52.25
& 55.12
& 59.88
& 70.00
& 60.63 \\

MiniCPM-o 4.5
& 59.50
& 69.69
& 58.47
& 64.60
& 52.70
& 88.75
& 52.50
& 53.50
& 52.62
& 60.50
& 65.00
& 80.25
& 61.87 \\

GLM-4.6V
& 57.38
& 63.09
& 51.95
& 57.03
& 49.80
& \underline{97.25}
& 52.12
& 50.88
& 51.38
& 52.25
& 53.25
& 69.25
& 57.92 \\

Qwen3-VL-235B-A22B-Instruct
& 67.63
& \underline{79.17}
& 69.77
& 72.15
& 65.90
& 97.00
& 57.88
& 54.75
& 56.38
& 69.00
& 66.88
& 93.63
& 70.06 \\

\midrule

\rowcolor{gray!12}
\multicolumn{14}{l}{
\textbf{\textit{MLLM-based Image Detectors}}
} \\

FakeVLM~\citep{wen2025fakevlm}
& 73.50
& 62.52
& \textbf{88.00}
& \textbf{99.39}
& 55.10
& 82.50
& 67.50
& 73.62
& \underline{83.38}
& 84.25
& 77.12
& 87.75
& 76.40 \\

UniGenDet~\citep{zhang2026unigendet}
& 80.25
& 63.74
& 81.05
& \underline{95.36}
& 44.65
& 86.50
& 70.50
& 68.62
& 73.25
& \underline{93.12}
& \underline{93.25}
& 93.25
& 78.03 \\

Veritas~\citep{tan2026veritas}
& 51.00
& 54.43
& 44.90
& 53.24
& 47.30
& 93.00
& 48.75
& 50.12
& 49.38
& 50.38
& 51.00
& 51.75
& 52.44 \\

\midrule

\rowcolor{blue!7}
\textbf{IVT-Guard (Ours)}
& \textbf{98.75}
& \textbf{93.32}
& \underline{87.02}
& 93.37
& 66.30
& \textbf{98.75}
& \textbf{90.63}
& \textbf{89.25}
& \textbf{89.63}
& \textbf{99.13}
& \textbf{92.88}
& \textbf{99.38}
& \textbf{92.66} \\

\bottomrule
\end{tabular}%
}
\end{table}

\subsection{Main Results}
\noindent\textbf{Comparison with State-of-the-Art Methods.}
As shown in \cref{tab:image_results,tab:video_results},
IVT-Guard achieves the highest average accuracy on both image and video detection, reaching 92.66\% and 86.82\%, respectively.
It delivers improvements of 9.57\% and 13.32\% over the strongest modality-specific baselines, PGC and DeMamba.
Even the substantially larger Qwen3-VL-235B-A22B-Instruct falls behind IVT-Guard,
suggesting that simply increasing the model size of generic MLLMs is insufficient to improve AIGC detection accuracy.
IVT-Guard also surpasses MLLM-based detectors such as UniGenDet and Skyra-RL by 14.63\% and 24.10\%, respectively,
while maintaining strong cross-dataset and OOD generalization across both modalities.
These results demonstrate the effectiveness of our three-stage training paradigm in jointly strengthening modality-specific artifact perception,
evidence-grounded reasoning, and detection performance.
Text detection results are reported in Appendix~\cref{tab:text_results}.


\subsection{Ablation Study}
\begin{figure}[!t]
  \centering
  \begin{minipage}[t]{0.49\textwidth}
    \vspace{0pt}
    \begin{minipage}[t]{\linewidth}
\centering
\captionof{table}{Ablation study of the training stages.
Results are reported as ID/OOD accuracy (\%).}
\label{tab:ablation_training_stages}
\resizebox{\linewidth}{!}{%
\begin{tabular}{lccc}
\toprule
\textbf{Method} & \textbf{Image} & \textbf{Video} & \textbf{Text} \\
\midrule
Base
& 65.25/65.71 & 52.12/48.70 & 43.25/45.75 \\

\midrule
\multicolumn{4}{l}{\cellcolor{gray!12}\textit{w/o Artifact-Aware Pre-training}} \\
+ SFT
& 91.00/85.96 & 57.50/54.18 & 98.38/97.46 \\
+ SFT + GRPO
& 89.75/87.15 & 69.13/63.00 & \textbf{98.88}/97.71 \\

\midrule
\multicolumn{4}{l}{\cellcolor{gray!12}\textit{w/ Artifact-Aware Pre-training}} \\
+ A2E-SFT
& \textbf{99.00}/92.27 & 91.88/78.40 & 98.75/97.63 \\
+ A2E-SFT + GRPO
& 98.63/92.00 & 91.25/81.88 & 98.75/97.87 \\
\rowcolor{blue!7}
+ A2E-SFT + EVC-GRPO
& 98.75/\textbf{93.48} & \textbf{92.00}/\textbf{82.80} & 98.38/\textbf{98.00} \\
\midrule
$\Delta$ vs Base (pp)
& \textcolor{red!75!black}{\textbf{+33.50/+27.77}}
& \textcolor{red!75!black}{\textbf{+39.88/+34.10}}
& \textcolor{red!75!black}{\textbf{+55.13/+52.25}} \\
\bottomrule
\end{tabular}%
}
\end{minipage}

    \par\vspace{-0.4em}
    \begin{minipage}[t]{\linewidth}
\centering
\captionof{table}{Ablation study of Perceptual Amplifiers and Artifact-Aware Injection in A2E-SFT.}
\label{tab:ablation_amplifier_injection}
\resizebox{\linewidth}{!}{%
\begin{tabular}{cclccc}
\toprule
\textbf{SPA} & \textbf{TPA} & \textbf{Fusion Setting} & \textbf{Image} & \textbf{Video} & \textbf{Text} \\
\midrule

\myxmark & \myxmark & None
& 91.00/85.96
& 57.50/54.18
& 98.38/97.46 \\

\midrule
\multicolumn{6}{l}{\cellcolor{gray!12}\textit{Perceptual Amplifiers Ablation}} \\
\mycmark & \myxmark & Global+Patch
& 98.00/91.79
& 59.13/53.25
& 97.25/97.21 \\

\myxmark & \mycmark & Global+Patch
& 91.38/85.75
& \textbf{92.88}/77.43
& 98.13/97.42 \\

\midrule
\multicolumn{6}{l}{\cellcolor{gray!12}\textit{Artifact-Aware Injection Ablation}} \\
\mycmark & \mycmark & Global-only
& 98.50/91.71
& 92.38/78.38
& 98.13/\textbf{97.83} \\

\mycmark & \mycmark & Patch-only
& 93.13/85.98
& 91.38/77.40
& 98.25/97.71 \\

\mycmark & \mycmark & Probe-attention
& 98.63/91.29
& 92.25/78.18
& 98.25/97.67 \\

\rowcolor{blue!7}
\mycmark & \mycmark & Global+Patch
& \textbf{99.00}/\textbf{92.27}
& 91.88/\textbf{78.40}
& \textbf{98.75}/97.63 \\


\bottomrule
\end{tabular}%
}
\end{minipage}

  \end{minipage}
  \begin{minipage}[t]{0.49\textwidth}
    \vspace{0pt}
    \centering
    \includegraphics[width=\linewidth]
      {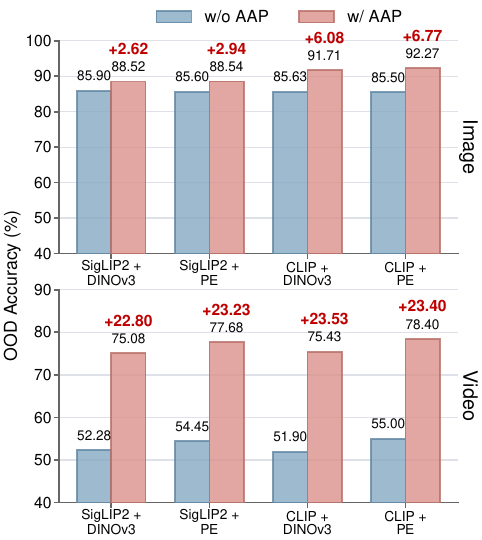}
    \captionsetup{aboveskip=-1pt, belowskip=0pt}
    \caption{Ablation study of visual encoder combinations in Artifact-Aware Pre-training.}
  \label{fig:encoder}
  \end{minipage}
\end{figure}

\noindent\textbf{Training-stage contributions.}
As shown in \cref{tab:ablation_training_stages},
standard SFT and GRPO yield limited improvements in detection performance, particularly for video.
Introducing AAP+A2E-SFT substantially alleviates this bottleneck,
improving image and video ID/OOD accuracy by 8.00\%/6.31\% and 34.38\%/24.22\% over standard SFT, respectively.
These results suggest that generic instruction tuning and reward optimization alone are insufficient to fully exploit forensic evidence,
demonstrating the benefit of integrating effective artifact representations into evidence-based reasoning.
Notably, vanilla GRPO after A2E-SFT yields only a slight gain in video OOD accuracy.
In contrast, EVC-GRPO consistently improves OOD performance across all three modalities,
suggesting that when artifact-aware perceptual representations are already sufficiently strong,
maintaining evidence--verdict consistency becomes crucial for further gains.

\begin{table}[!t]
\centering
\caption{
Accuracy comparison (\%) on the video modality of IVT-Set and three cross-dataset benchmarks.
$^{\dagger}$ denotes samples from four closed-source generators in AIGVDBench~\citep{ma2026AIGVDBench}.
\emph{Avg.} denotes the mean of ID accuracy, average CD accuracy, and average OOD accuracy.
Best and second-best results are shown in \textbf{bold} and \underline{underlined}, respectively.
}
\label{tab:video_results}

\resizebox{\textwidth}{!}{%
\begin{tabular}{
l
c
ccc
cccccc
c
}
\toprule
\multirow[c]{2}{*}{\textbf{Method}}
& \multirow[c]{2}{*}{\textbf{ID}}
& \multicolumn{3}{c}{\textbf{CD}}
& \multicolumn{6}{c}{\textbf{OOD}}
& \multirow[c]{2}{*}{\textbf{\emph{Avg.}}} \\
\cmidrule(lr){3-5}
\cmidrule(lr){6-11}
&
&
\textbf{LOKI}
& \textbf{AIGVDBench}$^{\dagger}$
& \textbf{GenBuster++}
& \textbf{Real}
& \textbf{EasyAnimateV5.1}
& \textbf{HunyuanVideo-I2V}
& \textbf{Kling 2.1}
& \textbf{Veo 3}
& \textbf{Vidu Q2}
& \\
\midrule

\rowcolor{gray!12}
\multicolumn{12}{l}{
\textbf{\textit{Video-specific Detectors}}
} \\

ReStraV~\citep{2025ReStraV}
& 69.25
& 36.64
& 44.79
& 52.00
& 77.25
& 48.62
& 67.62
& 58.63
& 50.62
& 54.37
& 57.75 \\

NSG-VD~\citep{zhang2025NSGVD}
& 50.38
& 57.46
& 78.73
& 50.40
& 18.75
& 46.00
& 45.25
& 50.12
& 48.00
& 47.88
& 51.75 \\

DeMamba~\citep{chen2026demamba}
& \underline{87.62}
& \underline{79.25}
& 44.22
& 62.45
& \underline{88.25}
& \underline{67.00}
& \underline{76.00}
& 63.88
& \underline{55.88}
& \underline{74.38}
& \underline{73.50} \\

\midrule

\rowcolor{gray!12}
\multicolumn{12}{l}{
\textbf{\textit{Generic MLLMs}}
} \\

Qwen2.5-VL-7B-Instruct
& 50.88
& 49.21
& 48.64
& 58.50
& 84.25
& 48.50
& 47.50
& 49.25
& 47.00
& 48.88
& 52.41 \\

Qwen3-VL-8B-Instruct
& 52.12
& 55.03
& 51.43
& 62.85
& 86.75
& 49.12
& 47.88
& 49.88
& 48.62
& 48.00
& 54.53 \\

InternVL3.5-8B
& 52.75
& 50.00
& 48.78
& 54.85
& 78.75
& 49.12
& 48.50
& 48.50
& 47.38
& 46.38
& 52.36 \\

MiniCPM-o 4.5
& 50.12
& 60.06
& 60.91
& 55.85
& 74.50
& 50.88
& 47.88
& 49.00
& 46.75
& 47.50
& 53.94 \\

GLM-4.6V
& 50.25
& 48.11
& 43.49
& 55.10
& 85.75
& 49.00
& 47.62
& 48.38
& 47.88
& 47.25
& 51.15 \\

Qwen3-VL-235B-A22B-Instruct
& 52.75
& 55.30
& 52.18
& 61.95
& 87.25
& 50.38
& 48.63
& 50.75
& 48.38
& 49.38
& 55.01 \\

\midrule

\rowcolor{gray!12}
\multicolumn{12}{l}{
\textbf{\textit{MLLM-based Video Detectors}}
} \\

BusterX++~\citep{wen2025busterx++}
& 52.12
& 66.77
& \underline{83.81}
& \textbf{83.20}
& 82.75
& 46.88
& 45.62
& 48.88
& 47.62
& 45.75
& 60.99 \\

Skyra-RL~\citep{li2026skyra}
& 60.00
& 56.66
& 80.76
& 51.40
& \textbf{97.25}
& 49.12
& 57.75
& \textbf{86.75}
& 49.12
& 51.38
& 62.72 \\


\midrule

\rowcolor{blue!7}
\textbf{IVT-Guard (Ours)}
& \textbf{92.00}
& \textbf{89.25}
& \textbf{88.64}
& \underline{77.60}
& 85.75
& \textbf{87.00}
& \textbf{85.25}
& \underline{79.63}
& \textbf{73.88}
& \textbf{88.25}
& \textbf{86.82} \\

\bottomrule
\end{tabular}%
}
\end{table}

\noindent\textbf{Complementarity of perceptual amplifiers.}
As shown in \cref{tab:ablation_amplifier_injection},
the Spatial Perceptual Amplifier (SPA) and Temporal Perceptual Amplifier (TPA) exhibit clear modality-specific strengths.
SPA alone improves image ID/OOD accuracy by 7.00\%/5.83\%,
whereas TPA alone improves video ID/OOD accuracy by 35.38\%/23.25\%.
Combining both amplifiers consistently improves OOD accuracy over either amplifier alone across all three modalities,
demonstrating the complementarity of spatial and temporal forensic representations.

\noindent\textbf{Design of Artifact-Aware Injection.}
In \cref{tab:ablation_amplifier_injection},
``Global'' denotes aggregated image and video features, while ``Patch'' denotes local image and frame-level video features.
``Probe-attention'' instead uses learnable probe queries to fuse artifact representations before injection.
The results show that Global-only already achieves strong performance, whereas Patch-only leads to a noticeable drop,
particularly for images, where Global+Patch improves ID/OOD accuracy by 5.87\%/6.29\% over Patch-only.
This indicates that aggregated artifact representations provide the primary holistic forensic context,
while local or spatial features serve more effectively as complementary fine-grained evidence.
Compared with Probe-attention, Global+Patch further improves image and video OOD accuracy by 0.98\% and 0.22\%, respectively,
showing that preserving both holistic and fine-grained artifact cues is more effective than probe-based fusion.

\noindent\textbf{Visual encoder combinations.}
As shown in \cref{fig:encoder},
regardless of the visual encoder pairing adopted in the AAP stage,
the subsequent A2E-SFT consistently achieves higher OOD accuracy than its counterpart without AAP,
with gains of 2.62\%--6.77\% on images and 22.80\%--23.53\% on videos.
These results indicate that AAP provides a backbone-agnostic mechanism for enhancing artifact-aware visual representations,
while preserving the modality-specific strengths of different encoders.
Among the combinations, CLIP+PE achieves the best overall detection performance.
More experimental results and analyses are provided in Appendix~\ref{apd:additional_experiments}.

\section{Conclusion}
In this work, we introduced IVT-Set and IVT-Guard for unified and interpretable AIGC detection across image, video, and text.
IVT-Set provides large-scale, multi-granularity Chain-of-Thought supervision for evidence-grounded detection reasoning.
Building on IVT-Set, IVT-Guard employs a three-stage training paradigm comprising Artifact-Aware Pre-training,
Artifact-to-Evidence Supervised Fine-Tuning, and Evidence-Verdict Consistency Group Relative Policy Optimization.
These stages respectively enhance modality-specific artifact perception, integrate artifact representations into evidence-grounded reasoning,
and enforce consistency between generated evidence and final verdicts, alleviating the reasoning--detection optimization dilemma.
This training paradigm mitigates the optimization dilemma between detection accuracy and reasoning quality.
Extensive experiments demonstrate that IVT-Guard consistently outperforms modality-specific detectors, generic MLLMs, and specialized MLLM-based detectors
across the evaluated modalities and generalization settings, while producing reliable and interpretable explanations.


\bibliography{iclr2027_conference}
\bibliographystyle{iclr2027_conference}

\clearpage
\appendix
\section{Appendix}
This appendix provides additional details on the implementation and training configurations,
supplementary quantitative experiments,
the construction of IVT-Set,
and Chain-of-Thought annotation prompts.
It is organized as follows:
\begin{enumerate}
    \renewcommand{\labelenumi}{$\bullet$~\textbf{A.\arabic{enumi}.}}
    \setlength{\itemsep}{0.15em}
    \setlength{\parskip}{0pt}
    \setlength{\parsep}{0pt}

    \item \textbf{More Implementation Details}
    \begin{enumerate}
        \renewcommand{\labelenumii}{$\bullet$~A.\arabic{enumi}.\arabic{enumii}.}
        \setlength{\itemsep}{0em}
        \setlength{\parskip}{0pt}
        \setlength{\parsep}{0pt}
        \item Training and Evaluation Datasets
        \item Comparison Baselines
        \item Training and Inference Details
    \end{enumerate}

    \item \textbf{Additional Quantitative Experiments}
    \begin{enumerate}
        \renewcommand{\labelenumii}{$\bullet$~A.\arabic{enumi}.\arabic{enumii}.}
        \setlength{\itemsep}{0em}
        \setlength{\parskip}{0pt}
        \setlength{\parsep}{0pt}
        \item Text Modality Detection Performance
        \item Reliability of Evidence-Grounded Explanations
        \item Cross-Modal Joint Training and Complementarity
        \item Robustness under Modality-Specific Perturbations
        \item Effectiveness of the Three-Stage Training Paradigm Across MLLMs
    \end{enumerate}

    \item \textbf{Details of the IVT-Set}
    \begin{enumerate}
        \renewcommand{\labelenumii}{$\bullet$~A.\arabic{enumi}.\arabic{enumii}.}
        \setlength{\itemsep}{0em}
        \setlength{\parskip}{0pt}
        \setlength{\parsep}{0pt}
        \item Generative Methods
        \item Detailed Statistics of IVT-Set
    \end{enumerate}

    \item \textbf{Chain-of-Thought Annotation Prompts}
\end{enumerate}
\subsection{More Implementation Details}
\label{apd:appendix_implementation_details}
\subsubsection{Training and Evaluation Datasets}
All training and evaluation data cover three modalities: image, video, and text.
Training and in-domain (ID) evaluation sets use a balanced 1:1 ratio between real and fake samples.
For out-of-domain (OOD) evaluation, each unseen generator is evaluated using
400 fake samples together with a shared set of 400 real samples.
OOD metrics are computed separately for each generator and then macro-averaged
across unseen generators.
Detailed in-domain (ID) and out-of-domain (OOD) splits for each modality are provided in~\cref{tab:image_gen_methods,tab:video_gen_methods,tab:text_gen_methods}.

\noindent\textbf{Training Data.}
We construct the training set with broad coverage of semantic categories and generator types.
The training set contains 45K samples across the three modalities, with 15K samples per modality,
including 10K for A2E-SFT and 5K for EVC-GRPO.

\noindent\textbf{Evaluation Data.}
\label{apd:evaluation_data}
For each modality, we construct an in-domain (ID) test set containing 400 real and 400 fake samples,
where the fake samples are generated by the same generator pool used for training, to evaluate within-distribution performance.
To further assess out-of-distribution (OOD) generalization, we construct an additional OOD test set for each modality,
consisting of a shared set of 400 real samples and 400 fake samples from each generator unseen during training.
We compute OOD metrics for each unseen generator separately and report their macro-average.
For cross-dataset (CD) evaluation, we use four image benchmarks (Chameleon~\citep{yan2025AIDE},
LOKI~\citep{ye2025loki}, GenImage~\citep{zhu2023genimage}, and GenBuster++~\citep{wen2025busterx++}),
three video benchmarks (LOKI, AIGVDBench~\citep{ma2026AIGVDBench}, and GenBuster++),
and four text benchmarks (LOKI, BiScope~\citep{guo2024biscope},
DetectRL-X~\citep{wu2026detectrl}, and AI Detector Bench~\citep{ai_detector_bench2026}).

\subsubsection{Comparison Baselines}
We compare IVT-Guard with modality-specific detectors, generic MLLMs/LLMs,
and MLLM/LLM-based detectors.

For image detection, modality-specific baselines include
UnivFD~\citep{ojha2023UnivFD}, DIRE~\citep{wang2023DIRE}, NPR~\citep{tan2024NPR},
AIDE~\citep{yan2025AIDE}, DGS-Net~\citep{yan2026dgs}, and PGC~\citep{zhou2026pgc}.
MLLM-based detectors include FakeVLM~\citep{wen2025fakevlm},
UniGenDet~\citep{zhang2026unigendet}, and Veritas~\citep{tan2026veritas}.
Generic MLLM baselines include Qwen2.5-VL-7B-Instruct~\citep{Qwen2.5-VL},
Qwen3-VL-8B-Instruct~\citep{Qwen3-VL},
InternVL3.5-8B~\citep{wang2025internvl3_5},
MiniCPM-o 4.5~\citep{cui2026minicpmo45},
GLM-4.6V~\citep{vteam2025glm45v},
and Qwen3-VL-235B-A22B-Instruct~\citep{Qwen3-VL}.

For video detection, modality-specific baselines include
ReStraV~\citep{2025ReStraV}, NSG-VD~\citep{zhang2025NSGVD},
and DeMamba~\citep{chen2026demamba}.
MLLM-based detectors include BusterX++~\citep{wen2025busterx++}
and Skyra-RL~\citep{li2026skyra}.
We use the same generic MLLM baselines as in image detection.

For text detection, modality-specific baselines include
ModernBERT~\citep{warner2025modernbert,drayson2025modernbert_detect}
and Desklib AI~\citep{2025Desklib}.
Generic LLM/MLLM baselines include Qwen3-VL-8B-Instruct,
InternVL3.5-8B, gpt-oss-20b~\citep{agarwal2025gptoss},
Qwen3-30B-A3B~\citep{qwen3}, and DeepSeek-V4-Flash~\citep{xu2026deepseek}.
LLM-based detectors include DivEye~\citep{diveye25},
FourierGPT~\citep{xu2024FourierGPT},
DetectAnyLLM~\citep{fu2025detectanyllm},
and WaveDetect~\citep{liu2026wavedetect}.

\subsubsection{Training and Inference Details}
\label{apd:appendix_training_inference_details}

For AAP, the Spatial Perceptual Amplifier (SPA) is initialized from
CLIP ViT-L/14@336px and adapted with rank-8 LoRA
($\alpha=16$, dropout $0.05$), while the Temporal Perceptual Amplifier (TPA)
is initialized from PE-Core-L14-336 and fine-tunes its last six blocks together
with the output pooler and binary classification head.
We uniformly sample eight frames from each video.
SPA and TPA are trained for 20 and 3 epochs with global batch sizes of 32 and 16,
respectively.
The learning rate is set to $1\times10^{-4}$ for SPA.
For TPA, we use $5\times10^{-6}$ for the PE encoder and
$5\times10^{-5}$ for the classification head.
Both branches use cosine learning-rate decay.

For A2E-SFT, we initialize the MLLM from Qwen3-VL-8B-Instruct
and freeze the CLIP and PE encoders.
We train the modality-specific MLP projectors and cross-attention modules
used for artifact-aware injection, together with all-linear LoRA adapters
of the MLLM.
Specifically, the image branch optimizes the global and local projectors,
while the video branch optimizes the temporal and spatial projectors.
The LoRA rank and scaling factor are set to $128$ and $256$, respectively.
We train for 3 epochs with a learning rate of $1\times10^{-5}$
and a global batch size of 128.

For EVC-GRPO, we initialize the policy from the merged A2E-SFT checkpoint
and keep the auxiliary CLIP and PE encoders frozen.
We jointly optimize the Qwen3-VL policy together with the modality-specific
MLP projectors and cross-attention modules used for artifact-aware injection.
We train for 1 epoch with a learning rate of $1\times10^{-6}$
and a global batch size of 128.
The PPO mini-batch size is 32, and eight rollouts are sampled per prompt.
The reward combines accuracy, format compliance, and evidence--verdict
consistency with weights $0.9$, $0.1$, and $0.2$, respectively,
without KL regularization.

For inference, we deploy IVT-Guard with vLLM~\citep{kwon2023vllm}.
For image and video inputs, the frozen auxiliary encoders extract artifact
representations, which are incorporated into the MLLM visual tokens through
the same artifact-aware injection mechanism used during training.
We use greedy decoding with temperature 0, top-$k=1$, top-$p=1.0$,
a repetition penalty of 1.0, and a maximum generation length of
4{,}096 tokens.

\subsection{Additional Quantitative Experiments}
\label{apd:additional_experiments}

\subsubsection{Text Modality Detection Performance}
\begin{table}[!t]
\centering
\caption{
Performance comparison (Acc, \%) on the text modality of IVT-Set
and three cross-dataset benchmarks.
Results are reported under in-domain (ID), cross-dataset (CD),
and out-of-domain (OOD) settings.
\emph{Avg.} denotes the average of the ID accuracy,
the mean CD accuracy, and the mean OOD accuracy.
Best results are in \textbf{bold} and the second-best are
\underline{underlined}.
}
\label{tab:text_results}

\resizebox{\textwidth}{!}{%
\begin{tabular}{
l
c
cccc
cccc
c
}
\toprule
\multirow[c]{2}{*}{\textbf{Method}}
& \multirow[c]{2}{*}{\textbf{ID}}
& \multicolumn{4}{c}{\textbf{CD}}
& \multicolumn{4}{c}{\textbf{OOD}}
& \multirow[c]{2}{*}{\textbf{\emph{Avg.}}} \\
\cmidrule(lr){3-6}
\cmidrule(lr){7-10}
&
&
\textbf{LOKI}
& \textbf{BiScope}
& \textbf{DetectRL-X}
& \textbf{AI Detector Bench}
& \textbf{Real}
& \textbf{Claude Opus 4.1}
& \textbf{Gemini 2.5 Pro}
& \textbf{GPT-4o}
& \\
\midrule

\rowcolor{gray!12}
\multicolumn{11}{l}{
\textbf{\textit{Text-specific Detectors}}
} \\

ModernBERT~\citep{warner2025modernbert,drayson2025modernbert_detect}
& 51.75
& 78.16
& 63.00
& 54.62
& 62.10
& 89.00
& 64.38
& 62.88
& 66.25
& 62.28 \\

Desklib AI~\citep{2025Desklib}
& 79.75
& \underline{93.75}
& \textbf{96.61}
& 84.28
& 83.40
& 91.00
& 93.50
& 91.62
& 91.88
& 87.09 \\

\midrule

\rowcolor{gray!12}
\multicolumn{11}{l}{
\textbf{\textit{Generic MLLMs \& LLMs}}
} \\

Qwen3-VL-8B
& 43.25
& 51.81
& 42.14
& 50.93
& 50.70
& 69.50
& 47.62
& 45.38
& 44.25
& 47.94 \\

InternVL3.5-8B
& 42.50
& 47.82
& 42.36
& 60.78
& 59.10
& 76.50
& 48.25
& 47.75
& 47.50
& 50.00 \\

gpt-oss-20b
& 44.38
& 66.46
& 68.94
& 49.64
& 55.90
& 44.75
& 48.75
& 45.00
& 46.12
& 50.26 \\

Qwen3-30B-A3B
& 36.88
& 63.76
& 61.51
& 52.75
& 51.60
& 48.25
& 41.75
& 40.75
& 40.25
& 45.68 \\

DeepSeek-V4-Flash
& 48.75
& 50.83
& 57.63
& 58.90
& 60.00
& 76.00
& 63.60
& 61.16
& 61.80
& 57.08 \\

\midrule

\rowcolor{gray!12}
\multicolumn{11}{l}{
\textbf{\textit{LLM-based Text Detectors}}
} \\

DivEye~\citep{diveye25}
& 71.25
& 54.38
& 57.33
& 60.53
& 58.80
& 69.00
& 67.38
& 61.88
& 64.75
& 64.92 \\

FourierGPT~\citep{xu2024FourierGPT}
& 54.00
& 62.54
& 55.62
& 56.07
& 54.80
& 47.75
& 48.88
& 52.00
& 52.88
& 53.88 \\

DetectAnyLLM~\citep{fu2025detectanyllm}
& 83.75
& 63.46
& 62.53
& 93.82
& \underline{89.40}
& \textbf{99.75}
& 87.75
& 84.62
& 89.75
& 83.84 \\

WaveDetect~\citep{liu2026wavedetect}
& \underline{94.25}
& 93.07
& \underline{95.18}
& \textbf{95.26}
& 87.00
& 92.25
& \underline{96.00}
& \underline{94.12}
& \underline{95.62}
& \underline{93.79} \\

\midrule

\rowcolor{blue!7}
\textbf{IVT-Guard (Ours)}
& \textbf{98.38}
& \textbf{94.97}
& 94.42
& \underline{94.39}
& \textbf{98.90}
& \underline{97.25}
& \textbf{97.88}
& \textbf{97.75}
& \textbf{98.38}
& \textbf{97.29} \\

\bottomrule
\end{tabular}%
}
\end{table}
As shown in~\cref{tab:text_results}, IVT-Guard achieves the highest overall average accuracy of 97.29\% on text detection,
outperforming the strongest text-specific baseline, WaveDetect, by 3.50 percentage points.
Generic MLLMs and LLMs achieve overall averages below 58\%,
suggesting that general-purpose language modeling capability alone is insufficient for reliable text authenticity detection.
IVT-Guard also surpasses LLM-based text detectors such as DetectAnyLLM by 13.45 percentage points,
while achieving 95.67\% average CD accuracy and 97.82\% average OOD accuracy.
These results demonstrate the effectiveness of IVT-Guard in jointly achieving accurate text authenticity detection
and strong generalization across datasets and unseen generators.

\subsubsection{Reliability of Evidence-Grounded Explanations}
\begin{table}[!t]
\centering
\caption{Explanation reliability evaluation on the OOD test set.
We report the absolute score evaluated by Qwen3.5-122B-A10B and
the pairwise Elo rating obtained from human evaluation for each modality,
with all methods initialized at an Elo rating of 1{,}000.}
\label{tab:elo_score_eval}
\begin{tabular}{lcccccc}
\toprule
\multirow{2}{*}{\textbf{Method}} & \multicolumn{2}{c}{\textbf{Image}} & \multicolumn{2}{c}{\textbf{Video}} & \multicolumn{2}{c}{\textbf{Text}} \\
\cmidrule(lr){2-3} \cmidrule(lr){4-5} \cmidrule(lr){6-7}
& Score & Elo & Score & Elo & Score & Elo \\
\midrule
GLM-4.6V & 3.09 & 1036 & 2.91 & 964 & 3.33 & 1132 \\
Qwen3-VL-235B-A22B & 3.88 & 1352 & 3.06 & 1024 & 3.68 & 1294 \\
Qwen3-VL-8B-Instruct & 3.91 & 1364 & 2.96 & 984 & 3.41 & 1164 \\
\rowcolor{blue!7}
IVT-Guard (Ours) & 4.70 & 1676 & 4.34 & 1564 & 4.28 & 1712 \\
\bottomrule
\end{tabular}%
\end{table}

As shown in~\cref{tab:elo_score_eval}, IVT-Guard achieves the highest explanation
scores and Elo ratings across all three modalities on the OOD test set.
Compared with the strongest baseline in each modality, it improves the
automatically evaluated explanation score by 0.60--1.28 points and the
human-evaluated Elo rating by 312, 540, and 418 points for image, video,
and text, respectively.
These results demonstrate that IVT-Guard provides higher-quality and more reliable
evidence-grounded explanations across modalities, beyond producing accurate
authenticity predictions.

\subsubsection{Cross-Modal Joint Training and Complementarity}
\begin{table}[!t]
\caption{Cross-modal evaluation and modality ablation at the A2E-SFT stage. Results are reported as ID/OOD accuracy and F1 (\%). Best results are in \textbf{bold}.}
\label{tab:cross_modal_transfer}
\resizebox{\linewidth}{!}{%
\begin{tabular}{lcccccc}
\toprule
\multirow{2}{*}{\textbf{Training Data}} & \multicolumn{3}{c}{\textbf{ID}} & \multicolumn{3}{c}{\textbf{OOD}} \\
\cmidrule(lr){2-4} \cmidrule(lr){5-7}
& Image (Acc/F1) & Video (Acc/F1) & Text (Acc/F1) & Image (Acc/F1) & Video (Acc/F1) & Text (Acc/F1) \\
\midrule
Image only       & 98.00/98.31 & 51.38/42.62 & 47.00/36.45 & 91.13/91.03 & 48.45/39.05 & 48.21/36.95 \\
Video only       & 68.00/67.54 & 91.25/91.25 & 45.38/39.43 & 66.73/63.79 & 75.73/75.14 & 46.54/39.80 \\
Text only        & 56.25/51.55 & 49.88/46.38 & 88.63/88.62 & 55.56/50.97 & 49.05/46.77 & 88.33/88.30 \\
Image + Video    & 98.75/98.75 & \textbf{92.63}/\textbf{92.62} & 46.00/39.73 & 91.58/91.52 & 77.23/76.63 & 46.88/39.25 \\
Video + Text     & 71.63/71.11 & 91.75/91.75 & 97.75/97.75 & 70.90/68.34 & \textbf{78.65}/\textbf{78.16} & 96.04/96.12 \\
Image + Text     & 98.63/98.75 & 50.50/38.80 & 98.13/98.12 & 92.13/92.09 & 48.98/37.02 & 96.96/96.96 \\
All       & \textbf{99.00}/\textbf{99.06} & 91.88/91.87 & \textbf{98.75}/\textbf{98.75} & \textbf{92.27}/\textbf{92.23} & 78.40/77.86 & \textbf{97.63}/\textbf{97.62} \\
\bottomrule
\end{tabular}%
}
\end{table}

As shown in~\cref{tab:cross_modal_transfer}, single-modality training exhibits a clear
diagonal pattern: each model performs strongly on its training modality but transfers
poorly to the other two modalities.
In contrast, joint training consistently benefits from complementary forensic cues.
Training on all three modalities achieves 99.00\%/92.27\%, 91.88\%/78.40\%,
and 98.75\%/97.63\% ID/OOD accuracy on image, video, and text, respectively,
improving OOD accuracy over the corresponding single-modality models by
1.14, 2.67, and 9.30 percentage points.
The gains are also asymmetric across modality combinations:
adding image data to text training improves text OOD accuracy from 88.33\% to 96.96\%,
whereas adding text data to image training increases image OOD accuracy only from
91.13\% to 92.13\%.
Overall, these results show that image, video, and text data can be jointly trained
within a unified model with little cross-modal interference, while benefiting from
complementary forensic information across modalities.

\subsubsection{Robustness under Modality-Specific Perturbations}
\begin{table}[!t]
\centering
\caption{
Robustness evaluation of IVT-Guard under modality-specific input perturbations.
Results are reported as accuracy/F1 (\%) under in-domain (ID) and
out-of-domain (OOD) settings for image, video, and text modalities.
}
\label{tab:comparison_perturbations}
\resizebox{\linewidth}{!}{%
\begin{tabular}{llcccccc}
\toprule
\multirow{2}{*}{\textbf{Method}} & \multirow{2}{*}{\textbf{Perturbation}} & \multicolumn{3}{c}{\textbf{ID}} & \multicolumn{3}{c}{\textbf{OOD}} \\
\cmidrule(lr){3-5} \cmidrule(lr){6-8}
& & Image (Acc/F1) & Video (Acc/F1) & Text (Acc/F1) & Image (Acc/F1) & Video (Acc/F1) & Text (Acc/F1) \\
\midrule
All & None & 98.75/98.75 & 92.00/91.99 & 98.38/98.37 & 93.48/93.43 & 82.80/82.71 & 98.00/98.00 \\
\midrule
\multirow{2}{*}{Image} & JPEG Compression (QF=90) & 98.38/98.37 & \textcolor{gray!30}{92.00/91.99} & \textcolor{gray!30}{98.38/98.37} & 93.69/93.64 & \textcolor{gray!30}{82.80/82.71} & \textcolor{gray!30}{98.00/98.00} \\
 & Gaussian Blur ($\sigma=1.0$) & 96.63/96.62 & \textcolor{gray!30}{92.00/91.99} & \textcolor{gray!30}{98.38/98.37} & 92.69/92.65 & \textcolor{gray!30}{82.80/82.71} & \textcolor{gray!30}{98.00/98.00} \\
 \midrule
\multirow{2}{*}{Video} & Shuffle frame & \textcolor{gray!30}{98.75/98.75} & 92.25/92.24 & \textcolor{gray!30}{98.38/98.37} & \textcolor{gray!30}{93.48/93.43} & 82.55/82.45 & \textcolor{gray!30}{98.00/98.00} \\
 & Resize $\times 0.7$ & \textcolor{gray!30}{98.75/98.75} & 89.13/89.06 & \textcolor{gray!30}{98.38/98.37} & \textcolor{gray!30}{93.48/93.43} & 81.93/81.87 & \textcolor{gray!30}{98.00/98.00} \\
 \midrule
\multirow{2}{*}{Text} & Human-like AI text & \textcolor{gray!30}{98.75/98.75} & \textcolor{gray!30}{92.00/91.99} & 98.38/98.37 & \textcolor{gray!30}{93.48/93.43} & \textcolor{gray!30}{82.80/82.71} & 98.08/98.08 \\
 & Adversarial prompt & \textcolor{gray!30}{98.75/98.75} & \textcolor{gray!30}{92.00/91.99} & 98.25/98.25 & \textcolor{gray!30}{93.48/93.43} & \textcolor{gray!30}{82.80/82.71} & 97.38/97.37 \\
\bottomrule
\end{tabular}%
}
\end{table}

As shown in~\cref{tab:comparison_perturbations}, IVT-Guard maintains strong
performance under a range of modality-specific input perturbations.
Most perturbations cause only minor changes in both ID and OOD accuracy,
with the largest degradation remaining below 3 percentage points.
F1 scores follow trends consistent with accuracy across all settings.
These results demonstrate that IVT-Guard is robust to common perturbations
across image, video, and text modalities.

\subsubsection{Effectiveness of the Three-Stage Training Paradigm Across MLLMs}
\begin{figure}[!t]
\centering
\includegraphics[width=\linewidth]{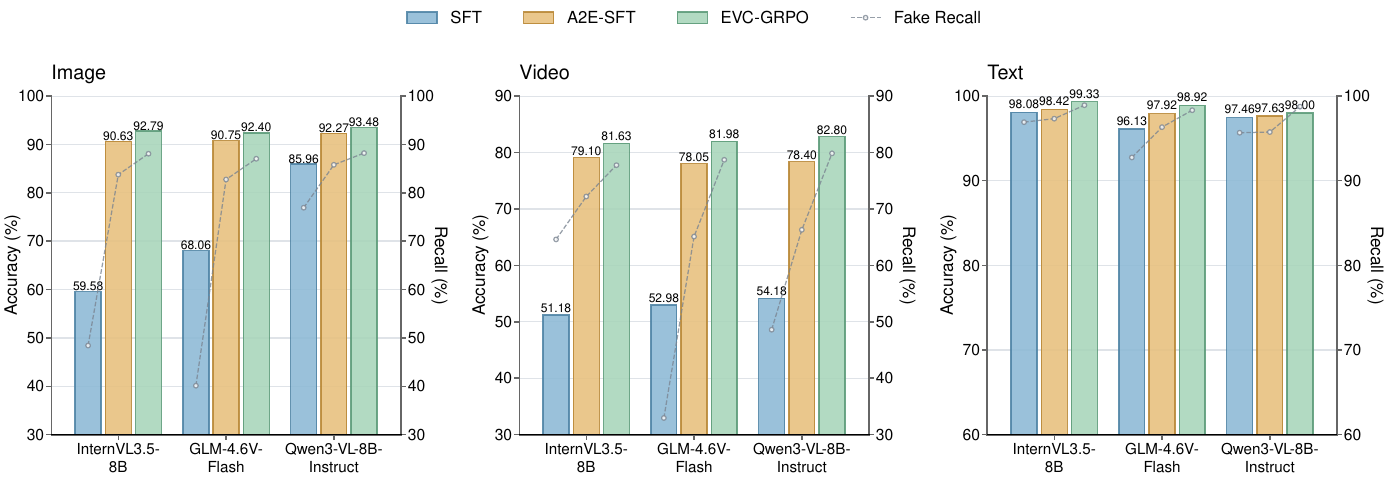}
\caption{
OOD accuracy across image, video, and text modalities for three model families at different training stages.
Bars denote OOD accuracy (\%), while lines denote mean recall (\%) across OOD generators,
with ``fake'' treated as the positive category.}
\label{fig:ood_stage_generalization}
\end{figure}
As shown in~\cref{fig:ood_stage_generalization}, the improvements introduced by
our training paradigm generalize consistently across different MLLM backbones.
A2E-SFT yields the largest gains, substantially improving OOD accuracy on both
image and video detection.
For example, InternVL3.5-8B improves from 59.58\% to 90.63\% on images
and from 51.18\% to 79.10\% on videos, while Qwen3-VL-8B-Instruct gains
6.31 and 24.22 percentage points, respectively.
EVC-GRPO further improves all three backbones, contributing an additional
1.21--2.16 points on images and 2.53--4.40 points on videos.
Text detection starts from a substantially stronger SFT baseline, yet still
benefits consistently from the subsequent stages.
Notably, after A2E-SFT and EVC-GRPO, the final OOD accuracies across the three
backbones differ by only 1.08 points on images and 1.17 points on videos,
despite their much larger gaps under standard SFT.
Together with the consistently increasing fake recall across training stages,
these results demonstrate that our three-stage paradigm provides robust gains
across different MLLM backbones and substantially reduces sensitivity to the
choice of base model when generalizing to unseen synthetic content.


\subsection{Details of the IVT-Set}
\begin{figure}[!t]
  \centering
  \includegraphics[width=\textwidth]{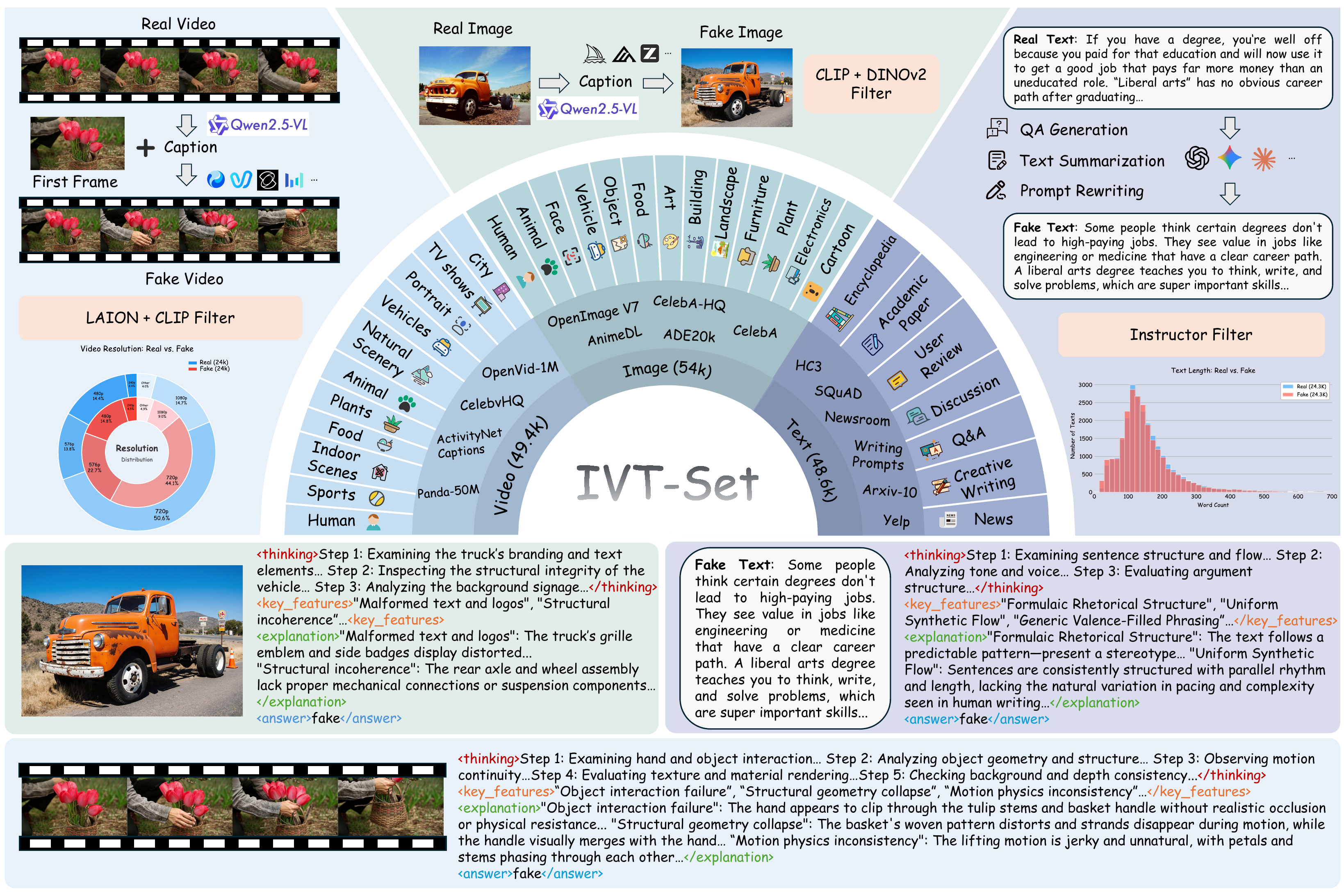}
  \caption{
Overview of the IVT-Set construction pipeline, including modality-specific
data pairing and filtering, diverse source and category coverage, and
multi-granularity CoT annotations.
}
  \label{fig:data_gen}
\end{figure}

The overall construction pipeline of IVT-Set is illustrated in~\cref{fig:data_gen}.
For each modality, we collect diverse real-world data across multiple sources and semantic categories,
construct corresponding synthetic samples using a broad range of generative models,
and apply modality-specific filtering to ensure data quality and reduce trivial real--fake discrepancies.
This process results in a large-scale multimodal benchmark comprising 54K images, 49.4K videos, and 48.6K texts.
We further annotate each sample with a unified multi-granularity reasoning structure,
consisting of \texttt{thinking}, \texttt{key\_features}, \texttt{explanation}, and \texttt{answer},
to connect modality-specific forensic cues with the final authenticity verdict.
We next describe the construction and statistics of each modality in detail.

\begin{table}[!t]
\centering
\caption{Details of the image modality generation methods in IVT-Set. The dataset includes 20 generators spanning GAN, diffusion/flow-based, and autoregressive paradigms, including a closed-source system whose architecture is not publicly disclosed. Generators are split into in-domain (ID) and out-of-domain (OOD) to evaluate cross-generator
generalization. Image sizes are reported as width $\times$ height in pixels. For generators with mixed-resolution samples, we report up to the four
most common image sizes in descending order.}
\label{tab:image_gen_methods}
\resizebox{\textwidth}{!}{
\begin{tabular}{lcccc}
\toprule
\textbf{Method} & \textbf{Model Type} & \makecell{\textbf{Resolution}} & \textbf{Data Scale} & \textbf{Data Split} \\
\midrule

BigGAN~\citep{brock2018biggan}
& GAN & 200$\times$200 / 128$\times$128 & 742 & ID \\

StyleGAN~\citep{karras2019stylegan}
& GAN & 200$\times$200 & 1,155 & ID \\

DF-GAN~\citep{tao2022dfgan}
& GAN & 256$\times$256 & 198 & ID \\

GALIP~\citep{tao2023galip}
& GAN & 224$\times$224 & 994 & ID \\

GigaGAN~\citep{kang2023gigagan}
& GAN & 512$\times$512 & 497 & ID \\

Stable Diffusion v1.4~\citep{compvis2024sd14}
& Diffusion & \makecell{1024$\times$704 / 1024$\times$768 \\ 704$\times$1024 / 768$\times$1024} & 767 & ID \\

SDXL 1.0 Base~\citep{podell2024sdxl}
& Diffusion & \makecell{1024$\times$704 / 1024$\times$768 \\ 768$\times$1024 / 704$\times$1024} & 1,127 & ID \\

PixArt-$\alpha$~\citep{chen2023pixartalpha}
& Diffusion & \makecell{1024$\times$704 / 1024$\times$768 \\ 704$\times$1024 / 768$\times$1024} & 881 & ID \\

Playground v2.5~\citep{li2024playground}
& Diffusion & \makecell{1024$\times$704 / 1024$\times$768 \\ 704$\times$1024 / 768$\times$1024} & 1,202 & ID \\

Stable Diffusion 3.5 Large~\citep{stabilityai2024sd35}
& Diffusion & \makecell{1024$\times$704 / 1024$\times$768 \\ 704$\times$1024 / 768$\times$1024} & 1,903 & ID \\

Stable Diffusion 2.1~\citep{rombach2021stable_diffusion}
& Diffusion & \makecell{1024$\times$704 / 1024$\times$768 \\ 704$\times$1024 / 768$\times$1024} & 1,038 & ID \\

Animagine XL 4.0~\citep{cagliostrolab2025animaginexl40}
& Diffusion & \makecell{704$\times$1024 / 704$\times$960 \\ 576$\times$768 / 640$\times$896} & 235 & ID \\

CogView4-6B~\citep{zheng2024cogview3}
& Diffusion & \makecell{1024$\times$704 / 1024$\times$768 \\ 704$\times$1024 / 768$\times$1024} & 2,584 & ID \\

Lumina-Image 2.0~\citep{lumina2}
& Diffusion & \makecell{1024$\times$704 / 1024$\times$768 \\ 704$\times$1024 / 768$\times$1024} & 2,139 & ID \\

Midjourney V6~\citep{midjourney}
& Closed-source & \makecell{1024$\times$771 / 771$\times$1024 \\ 1024$\times$1024 / 1024$\times$573} & 2,565 & OOD \\

FLUX.1 Krea [dev]~\citep{flux1kreadev2025}
& Diffusion & \makecell{1024$\times$704 / 1024$\times$768 \\ 704$\times$1024 / 768$\times$1024} & 2,368 & OOD \\

Qwen-Image~\citep{wu2025qwenimage}
& Diffusion & \makecell{1024$\times$704 / 1024$\times$768 \\ 704$\times$1024 / 768$\times$1024} & 2,525 & OOD \\

LlamaGen~\citep{sun2024Llamagen}
& Autoregressive & 512$\times$512 & 497 & OOD \\

Infinity~\citep{han2025infinity}
& Autoregressive & 512$\times$512 & 2,654 & OOD \\

Janus-Pro-7B~\citep{chen2025janus-pro}
& Autoregressive & 384$\times$384 & 2,749 & OOD \\

\midrule

\multicolumn{3}{c}{\textbf{Total: 20 generators}}
& \textbf{28,820} & \\

\bottomrule
\end{tabular}
}
\end{table}

\begin{table}[!t]
\centering
\caption{Details of the video modality generation methods in IVT-Set. 
The dataset includes 14 generators with diverse resolutions and frame rates. 
Generators are split into in-domain (ID) and out-of-domain (OOD) categories when applicable to evaluate cross-generator generalization. Video resolutions are reported as output heights in pixels. For generators with mixed-resolution samples, we report up to the four most common output heights in descending order.}
\label{tab:video_gen_methods}
\resizebox{\textwidth}{!}{
\begin{tabular}{lccccc}
\toprule
\textbf{Method} & \textbf{Duration (s)} & \makecell{\textbf{Resolution}} & \textbf{FPS} & \textbf{Data Scale} & \textbf{Data Split} \\
\midrule

Pika~\citep{pika2022}
& 3.00--4.00 & \makecell{768p / 576p \\ 640p / 1024p} & 12--24 & 1,000 & -- \\

SVD~\citep{blattmann2023svd}
& 3.57 & 576p & 7 & 1,000 & -- \\

DynamiCrafter~\citep{xing2024dynamicrafter}
& 2.00 & 576p & 8 & 1,000 & -- \\

OpenSora~\citep{opensora}
& 2.00 & 256p / 512p & 8 & 1,000 & -- \\

CogVideoX~\citep{yang2025cogvideox}
& 5.04 & 1024p & 24 & 1,000 & -- \\

LTX-Video~\citep{HaCohen2024LTXVideo}
& 1.37--10.25 & \makecell{672p / 480p \\ 512p / 240p} & 6--60 & 4,999 & ID \\

Hailuo 02~\citep{hailuo02_2025}
& 5.88--10.12 & \makecell{720p / 1080p \\ 512p / 480p} & 24 & 988 & ID \\

Wan 2.2~\citep{wan2025}
& 1.92--9.92 & \makecell{704p / 512p \\ 480p / 928p} & 6--60 & 3,999 & ID \\

Seedance 1.0~\citep{gao2025seedance}
& 5.04 & \makecell{704p / 1080p \\ 512p / 480p} & 24 & 990 & ID \\

HunyuanVideo-I2V~\citep{kong2024hunyuanvideo}
& 1.93--5.57 & \makecell{720p / 768p \\ 448p / 416p} & 23--60 & 2,000 & OOD \\

Vidu Q2~\citep{viduq2}
& 2.08--8.08 & \makecell{720p / 1080p \\ 512p / 480p} & 24 & 990 & OOD \\

EasyAnimateV5.1~\citep{xu2025easyanimate}
& 0.80--6.00 & \makecell{576p / 512p \\ 480p / 240p} & 7.5--60 & 3,998 & OOD \\

Kling 2.1~\citep{klingai}
& 5.04--10.04 & \makecell{720p / 512p \\ 480p / 960p} & 24 & 965 & OOD \\

Veo 3~\citep{veo3}
& 4.00 & \makecell{720p / 512p \\ 480p / 240p} & 24 & 806 & OOD \\

\midrule

\multicolumn{4}{c}{\textbf{Total: 14 generators}}
& \textbf{24,735} & \\

\bottomrule
\end{tabular}
}
\end{table}

\begin{table}[!t]
\centering
\caption{Details of the text modality generation methods in IVT-Set. 
The dataset includes 10 representative LLMs covering both open-source and closed-source models. 
Models are split into in-domain (ID) and out-of-domain (OOD) categories to evaluate cross-model generalization. 
R./F. Ratio denotes the total real-text length divided by the total corresponding generated-text length. Length reports the minimum--maximum generated-text length in words.}
\label{tab:text_gen_methods}
\resizebox{\textwidth}{!}{
\begin{tabular}{lcccc}
\toprule
\textbf{Method} & \textbf{R./F. Ratio} & \textbf{Length} & \textbf{Data Scale} & \textbf{Data Split} \\
\midrule

Llama-3.3-70B-Instruct~\citep{huang2024llama3}
& 1.09 & 20--1,190 & 1,486 & ID \\

gpt-oss-120b~\citep{agarwal2025gptoss}
& 0.99 & 27--1,056 & 2,635 & ID \\

GLM-4.5-Air~\citep{5team2025GLM-4.5}
& 1.04 & 20--1,116 & 2,469 & ID \\

DeepSeek-V3.1~\citep{liu2024deepseek}
& 1.09 & 20--1,516 & 2,473 & ID \\

Kimi K2-Instruct-0905~\citep{kimiteam2025kimik2}
& 1.05 & 20--787 & 2,190 & ID \\

Qwen3-235B-A22B-Instruct-2507~\citep{qwen3}
& 0.99 & 21--1,206 & 2,680 & ID \\

Qwen3-Next-80B-A3B-Instruct~\citep{qwen3}
& 1.04 & 20--1,270 & 2,451 & ID \\

GPT-4o~\citep{gpt4o}
& 0.99 & 20--1,098 & 2,500 & OOD \\

Claude Opus 4.1~\citep{claudeopus4}
& 0.98 & 20--1,528 & 2,730 & OOD \\

Gemini 2.5 Pro~\citep{gemini25}
& 1.03 & 24--1,033 & 2,722 & OOD \\

\midrule

\multicolumn{2}{c}{\textbf{Total: 10 LLMs}}
& & \textbf{24,336} & \\

\bottomrule
\end{tabular}
}
\end{table}

\subsubsection{Generative Methods}
To construct a comprehensive and challenging benchmark, we collect AI-generated samples from a diverse set of contemporary generators across image, video, and text modalities.
The selected generators span multiple architectural paradigms, release periods, and access types (open-source vs.\ commercial) to reduce bias toward specific generator families.
Generators with available split annotations are divided into in-domain (ID) and out-of-domain (OOD) splits to support the evaluation of cross-generator generalization.

\noindent\textbf{Image.}
As shown in~\cref{tab:image_gen_methods}, the image subset contains 28,820 generated images from 20 generators spanning GAN, diffusion/flow-based, and autoregressive paradigms, including a closed-source system whose architecture is not publicly disclosed.
(1) GAN-based models include early large-scale class-conditional generation with BigGAN~\citep{brock2018biggan}, style-disentangled synthesis with StyleGAN~\citep{karras2019stylegan}, and text-guided GANs such as DF-GAN~\citep{tao2022dfgan},
GALIP~\citep{tao2023galip}, and the scalable GigaGAN~\citep{kang2023gigagan}.
(2) Diffusion/flow-based models cover the evolution of modern generators, from Stable Diffusion v1.4~\citep{compvis2024sd14} to more recent models such as CogView4-6B~\citep{zheng2024cogview3}, Lumina-Image 2.0~\citep{lumina2}, Qwen-Image~\citep{wu2025qwenimage}, and FLUX.1 Krea [dev]~\citep{flux1kreadev2025}.
These models vary substantially in architectural design, model scale, training data, and supported resolutions.
(3) Closed-source systems with architectures that are not publicly disclosed include Midjourney V6~\citep{midjourney}.
(4) Autoregressive models include LlamaGen~\citep{sun2024Llamagen}, which extends LLM-style next-token prediction to image synthesis, Infinity~\citep{han2025infinity}, which adopts bitwise-parallel generation,
and the unified vision-language model Janus-Pro-7B~\citep{chen2025janus-pro}.
All autoregressive generators are assigned to the OOD split, yielding a challenging cross-model generalization setting.

\noindent\textbf{Video.}
As shown in~\cref{tab:video_gen_methods}, the video subset comprises 24,735 generated videos from 14 generators, with resolutions ranging from 240p to 1080p and frame rates from 6 to 60 FPS.
The collection covers early diffusion-based video models such as SVD~\citep{blattmann2023svd} and DynamiCrafter~\citep{xing2024dynamicrafter}, 
intermediate open-source systems including OpenSora~\citep{opensora} and CogVideoX~\citep{yang2025cogvideox},
and more recent high-fidelity generators such as HunyuanVideo-I2V~\citep{kong2024hunyuanvideo}, Wan 2.2~\citep{wan2025}, Seedance 1.0~\citep{gao2025seedance}, and Veo 3~\citep{veo3}.
We also include commercial platforms (Pika~\citep{pika2022}, Hailuo 02~\citep{hailuo02_2025}, Kling 2.1~\citep{klingai}, and Vidu Q2~\citep{viduq2}) that are widely used in real-world content creation, 
further enhancing the diversity and practical relevance of the benchmark.
For generators inherited from existing public benchmarks, we did not perform any split annotations.
For newly collected models, we assign ID/OOD labels to support systematic cross-source evaluation.

\noindent\textbf{Text.}
As shown in~\cref{tab:text_gen_methods}, the text subset comprises 24,336 machine-generated texts produced by 10 representative LLMs.
The selected models cover both open-source systems (Llama-3.3-70B-Instruct~\citep{huang2024llama3}, gpt-oss-120b~\citep{agarwal2025gptoss}, 
GLM-4.5-Air~\citep{5team2025GLM-4.5}, DeepSeek-V3.1~\citep{liu2024deepseek}, Kimi K2-Instruct-0905~\citep{kimiteam2025kimik2}, 
Qwen3-235B-A22B-Instruct-2507~\citep{qwen3}, and Qwen3-Next-80B-A3B-Instruct~\citep{qwen3}) and leading closed-source models (GPT-4o~\citep{gpt4o}, 
Claude Opus 4.1~\citep{claudeopus4}, and Gemini 2.5 Pro~\citep{gemini25}).
These models vary substantially in parameter scale, training strategy, and generation style, with fake text lengths ranging from 20 to 1,528 words.
The near-unity real-to-fake length ratios (0.98--1.09) limit overall length discrepancies between real and fake texts, reducing the risk that detectors exploit text length as a superficial cue.
Each generator contributes 1,486--2,730 samples, ensuring broad coverage without the benchmark being dominated by a single model.
We designate three closed-source models as OOD to evaluate whether detectors trained on open-source LLM outputs can generalize to unseen proprietary systems.

\begin{figure}[!t]
  \centering
  \begin{subfigure}[t]{0.48\textwidth}
    \centering
    \includegraphics[width=\textwidth]{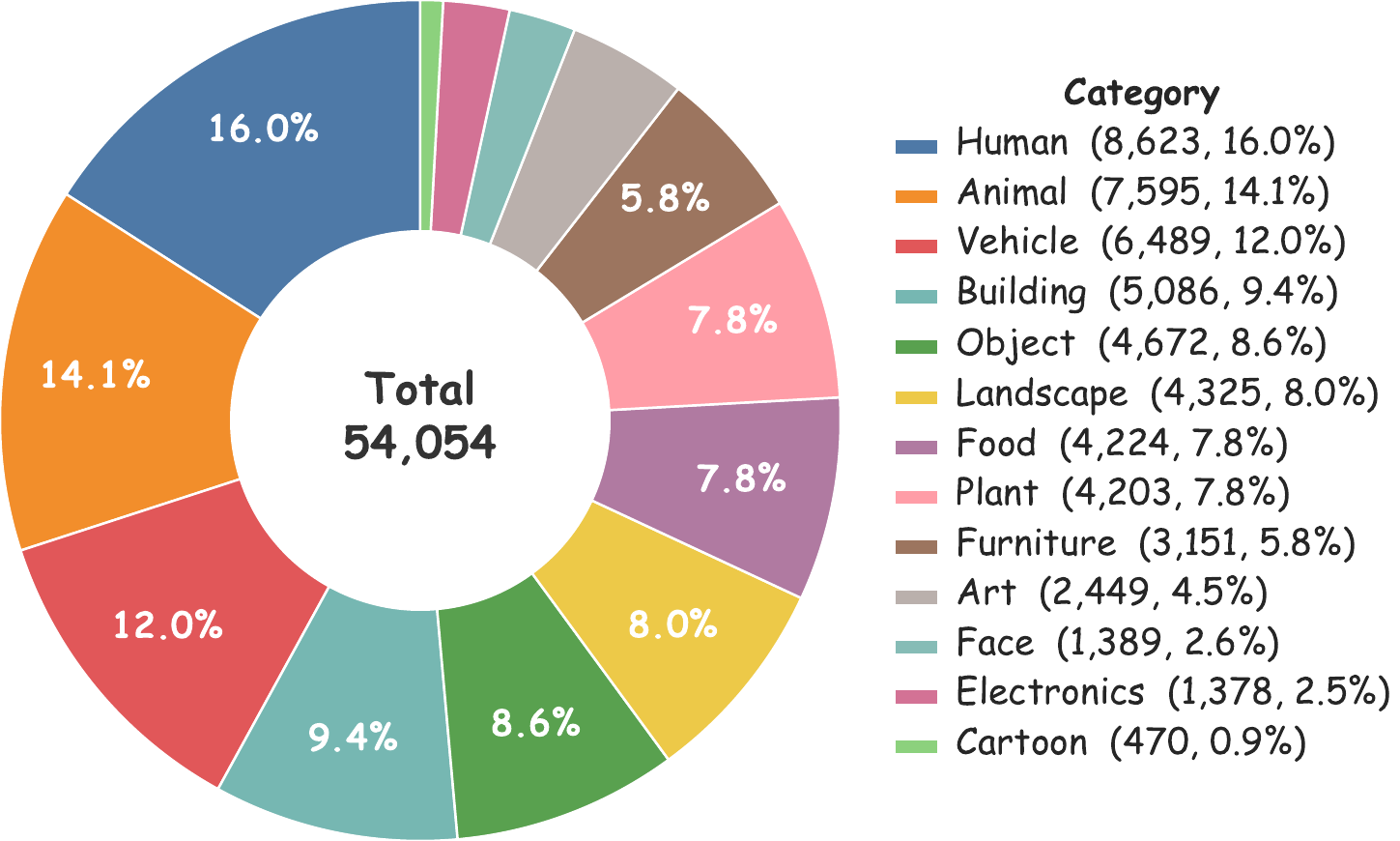}
    \caption{Category distribution of all images.}
    \label{fig:img_category_dist}
  \end{subfigure}
  \hfill
  \begin{subfigure}[t]{0.47\textwidth}
    \centering
    \includegraphics[width=\textwidth]{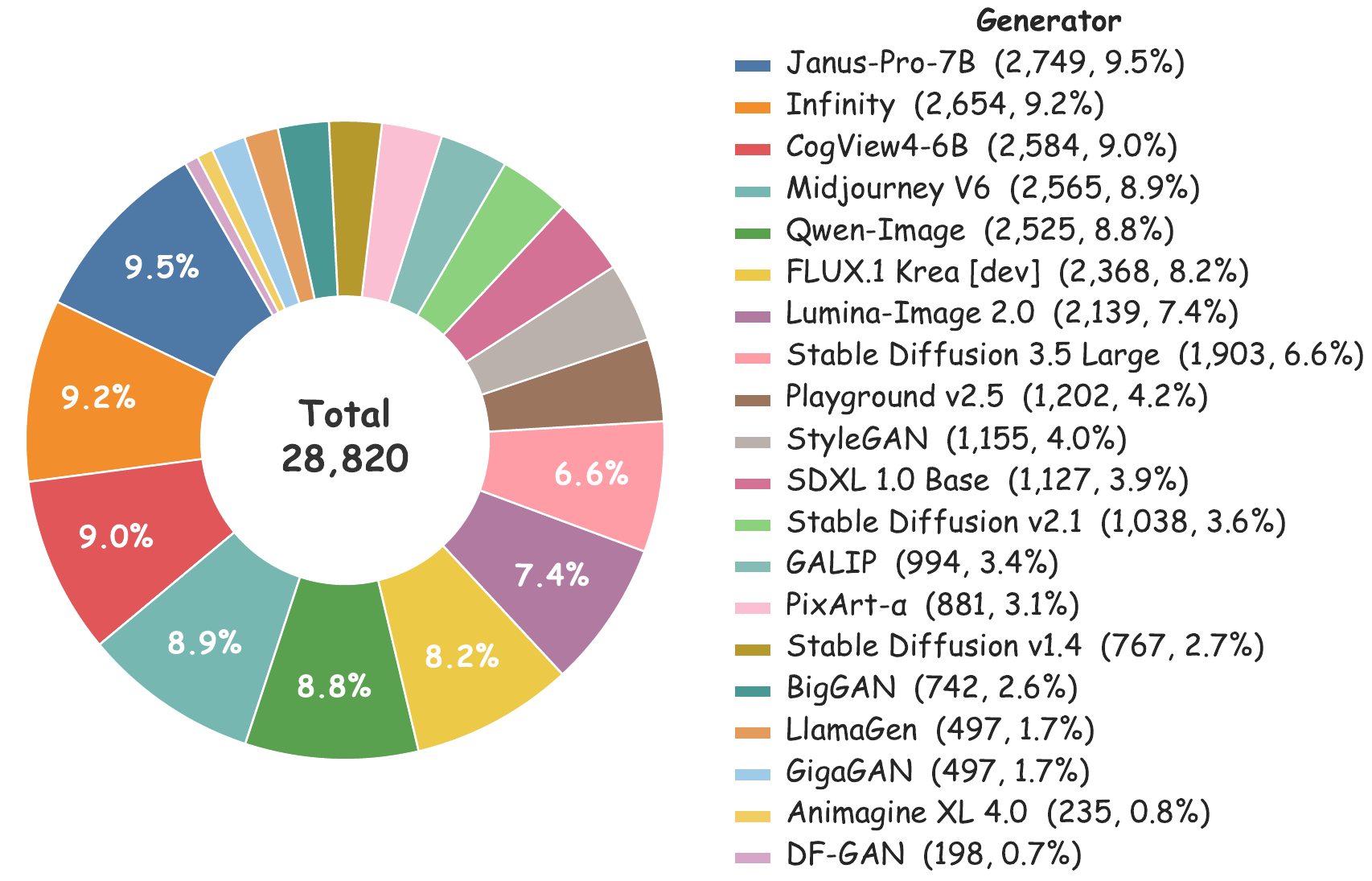}
    \caption{Generator distribution of fake images.}
    \label{fig:img_gen_dist}
  \end{subfigure}
  \begin{subfigure}[t]{\textwidth}
    \centering
    \includegraphics[width=\textwidth]{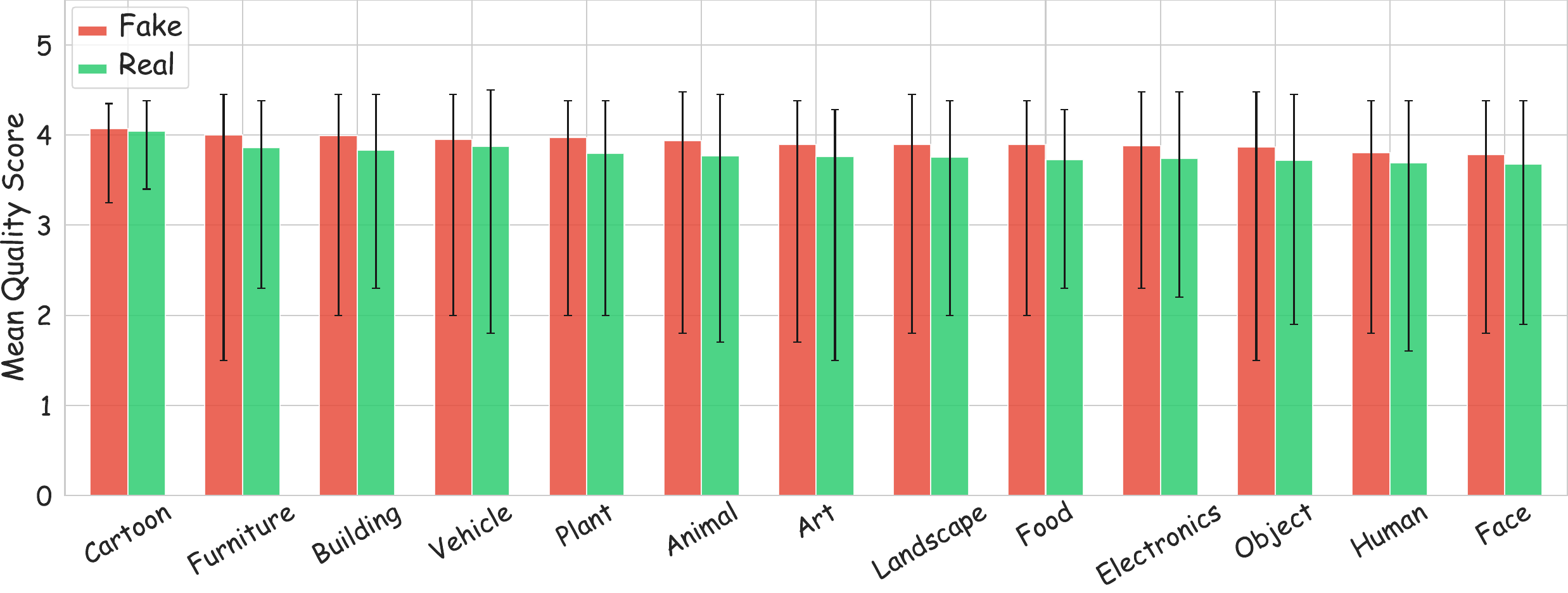}
    \caption{Comparison of the mean quality scores of fake and real images in each category.}
    \label{fig:img_quality_score}
  \end{subfigure}
  \caption{Data statistics of the image modality in IVT-Set. (a) Category distribution of all images across 13 semantic categories. (b) Generator distribution of fake images across 20 models spanning GAN, Diffusion, and Autoregressive paradigms. (c) Comparison of the mean quality scores of fake and real images in each category, showing that the quality scores of the two are very close across all categories.}
  \label{fig:data_stat_image}
\end{figure}
\begin{figure}[!t]
  \centering
  \begin{subfigure}[t]{0.32\textwidth}
    \centering
    \includegraphics[width=\textwidth]{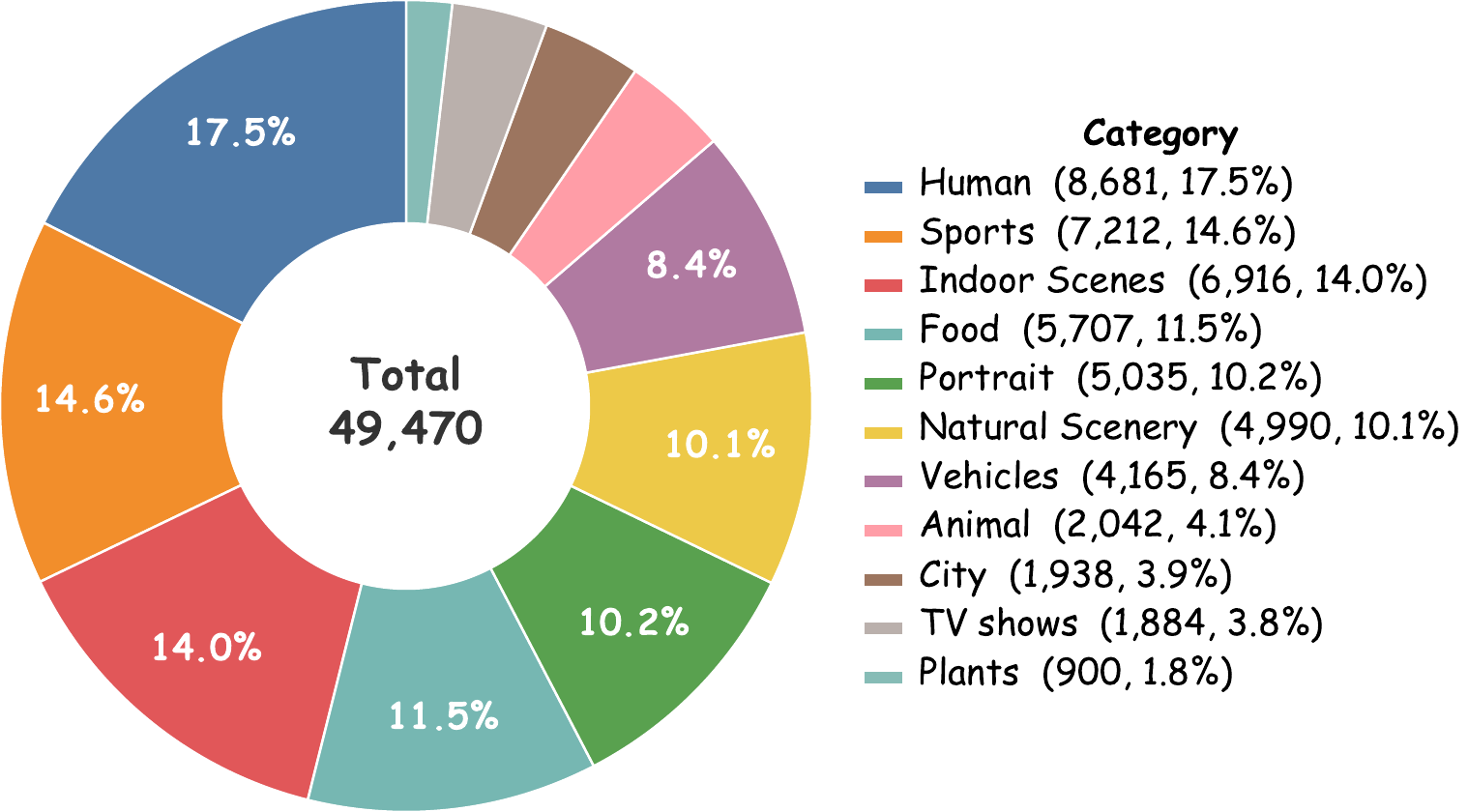}
    \caption{Category distribution of all videos.}
    \label{fig:vid_category_dist}
  \end{subfigure}
  \begin{subfigure}[t]{0.32\textwidth}
    \centering
    \includegraphics[width=\textwidth]{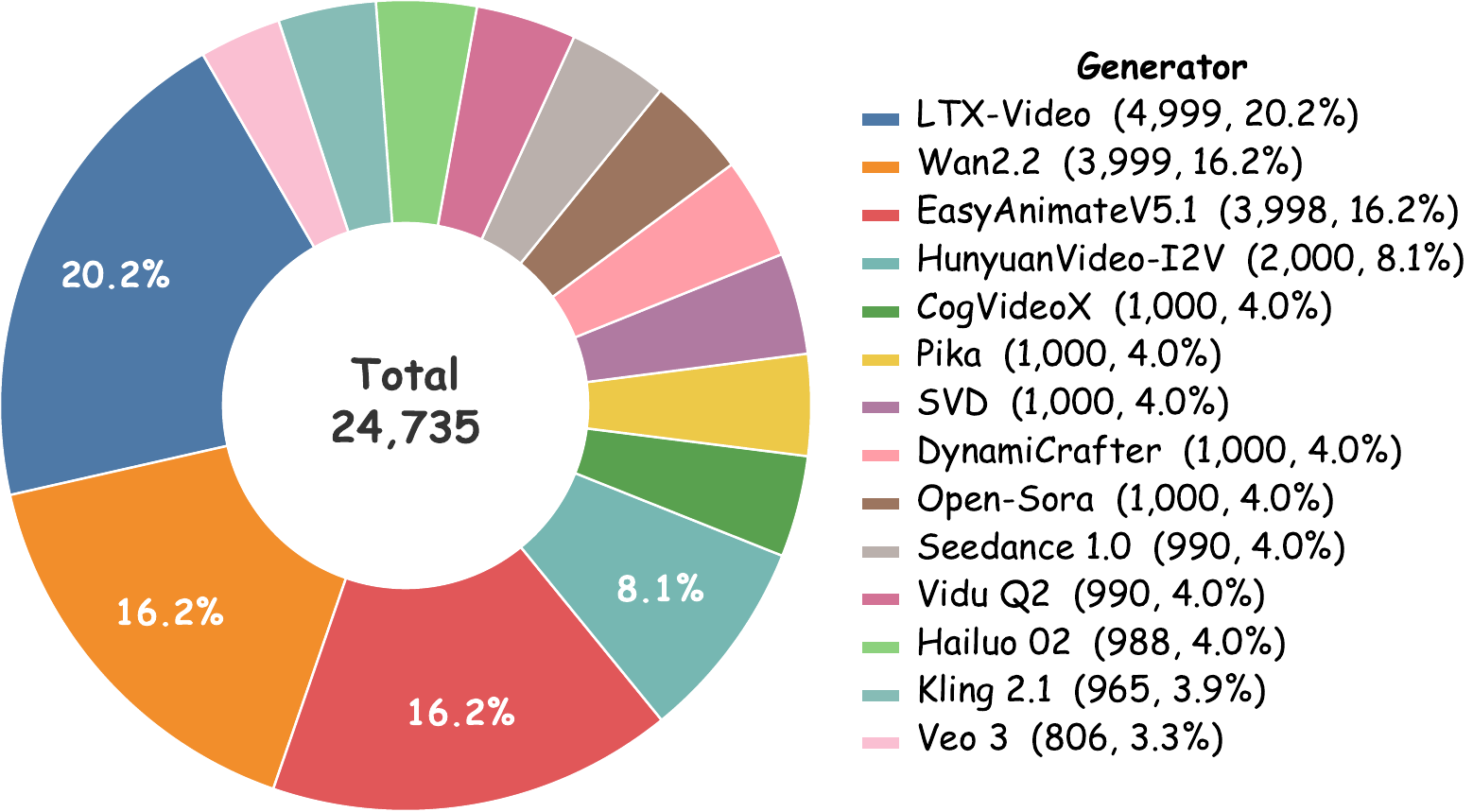}
    \caption{Generator distribution of fake videos.}
    \label{fig:vid_gen_dist}
  \end{subfigure}
  \begin{subfigure}[t]{0.3\textwidth}
    \centering
    \includegraphics[width=\textwidth]{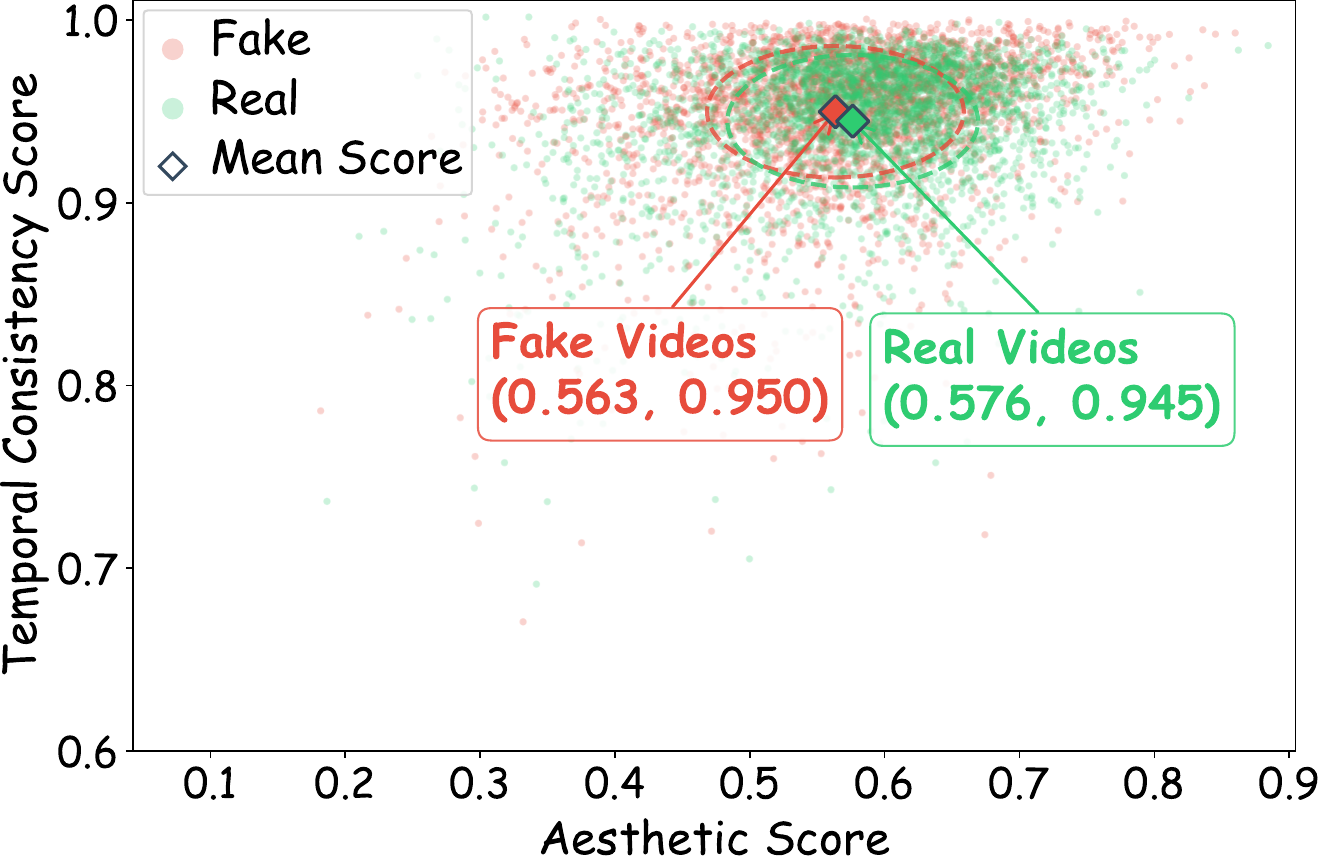}
    \caption{Correlation between aesthetic score and temporal consistency score.}
    \label{fig:vid_quality_score}
  \end{subfigure}
  \caption{Data statistics of the video modality in IVT-Set. (a) Category distribution of all videos across 11 semantic categories. (b) Generator distribution of fake videos across 14 generation models, covering both open-source and commercial models. (c) Correlation between aesthetic score and temporal consistency score for fake and real videos, showing that the two distributions largely overlap.}
  \label{fig:data_stat_video}
\end{figure}
\begin{figure}[!t]
  \centering
  \begin{subfigure}[t]{0.47\textwidth}
    \centering
    \includegraphics[height=0.46\textwidth]{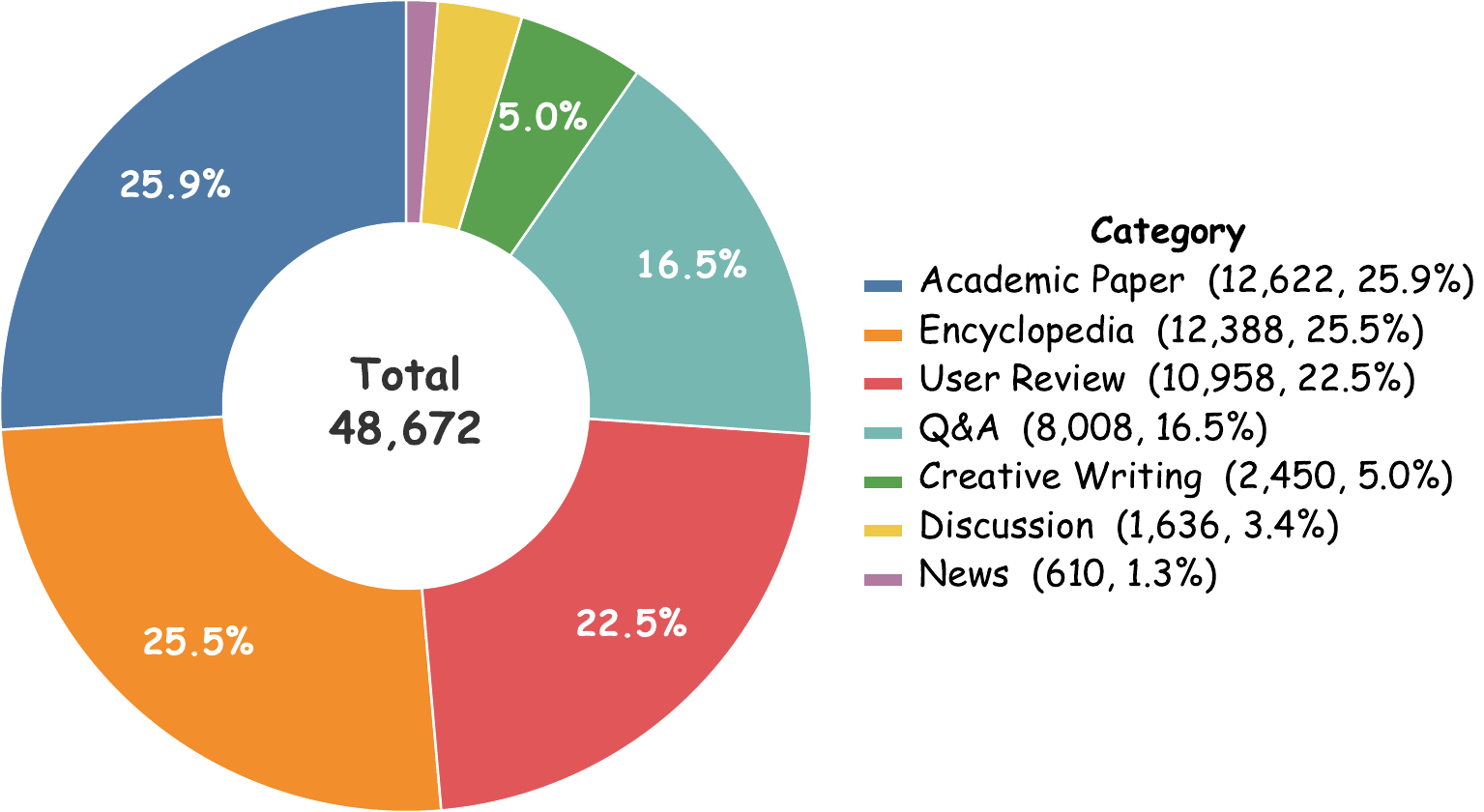}
    \caption{Category distribution of all texts.}
    \label{fig:txt_category_dist}
  \end{subfigure}
  \hfill
  \begin{subfigure}[t]{0.47\textwidth}
    \centering
    \includegraphics[height=0.46\textwidth]{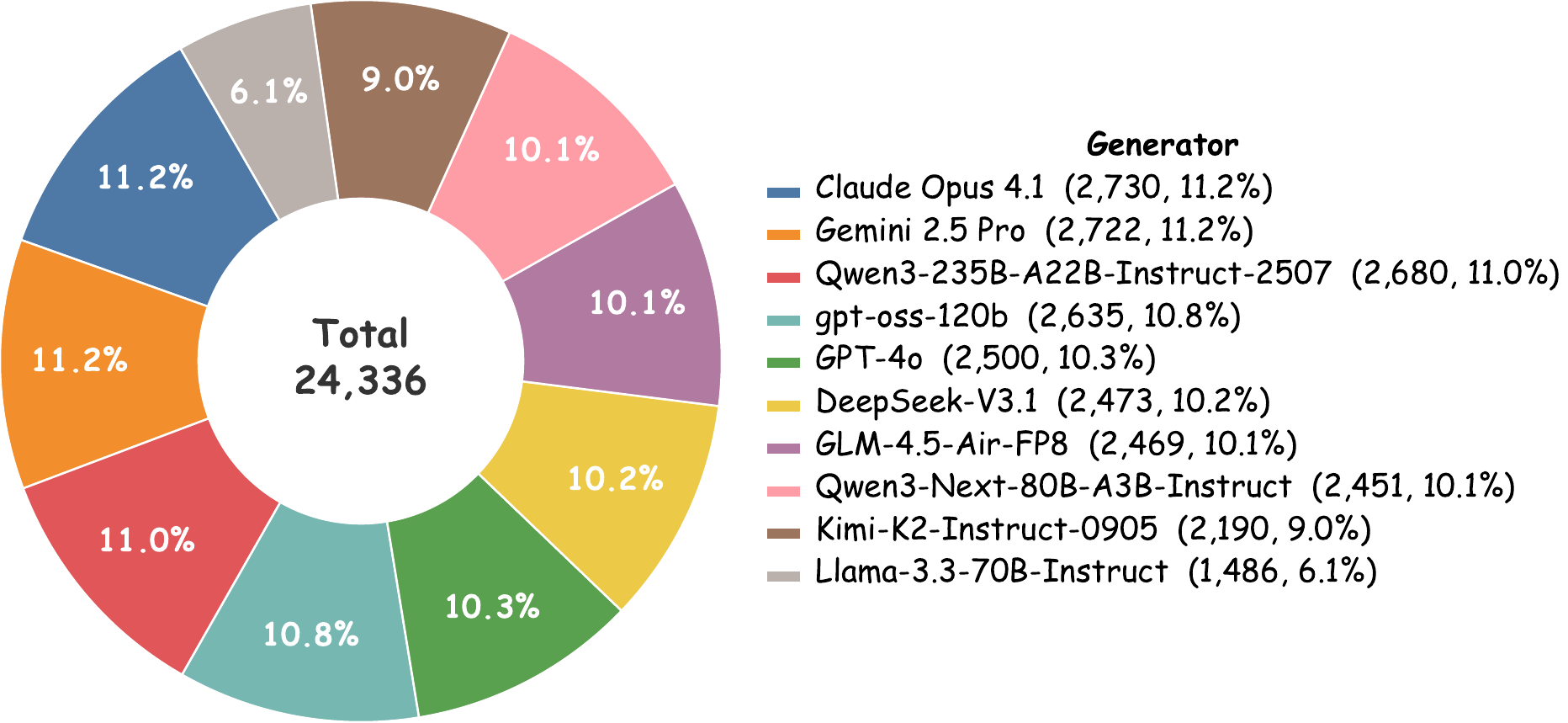}
    \caption{Generator distribution of fake texts.}
    \label{fig:txt_gen_dist}
  \end{subfigure}
  \par\vspace{0.6em}
  \begin{subfigure}[t]{0.4\textwidth}
    \includegraphics[width=\textwidth]{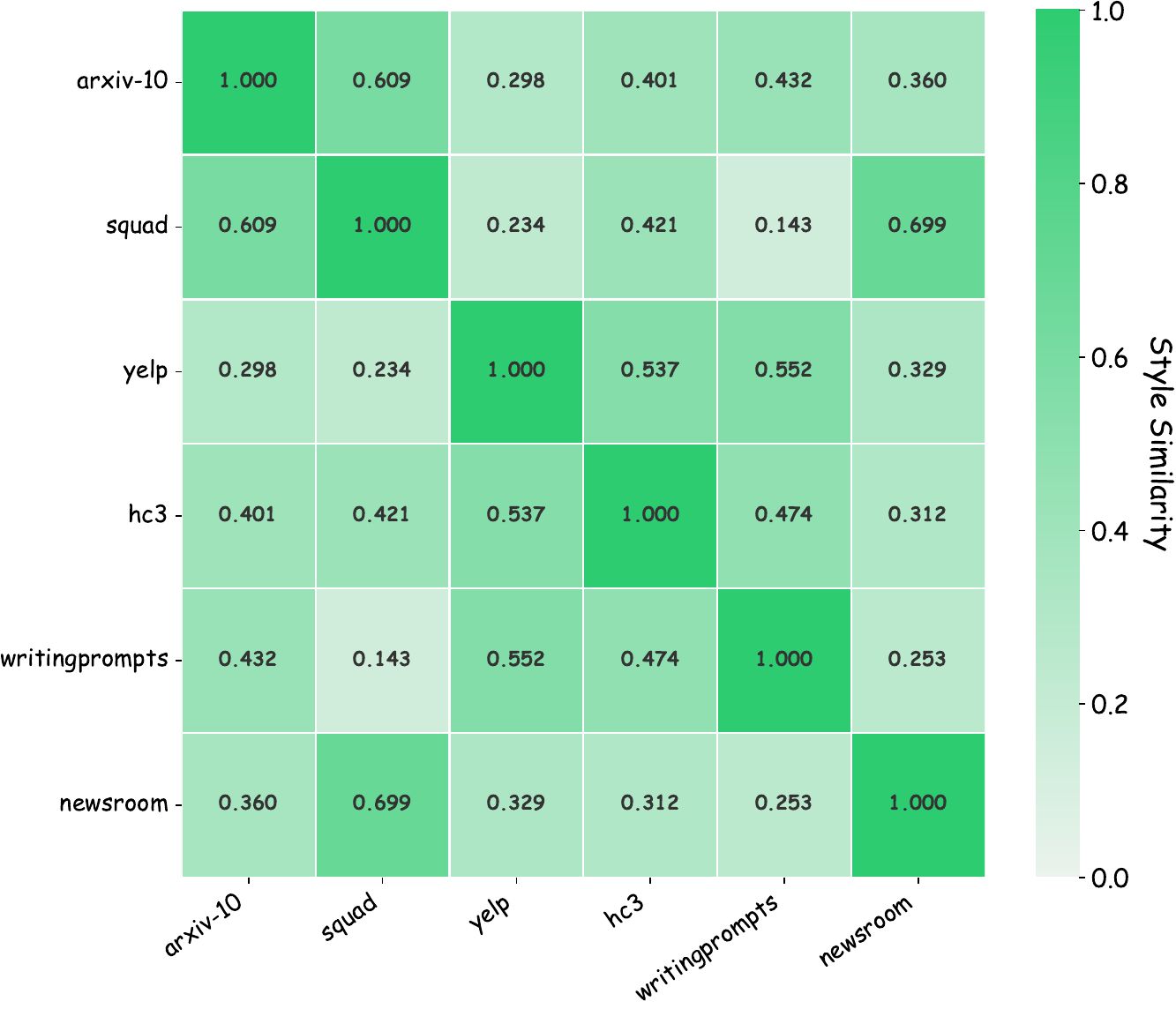}%
    \caption{Stylometric similarity among real sources.}
    \label{fig:txt_real_source_style_similarity}
  \end{subfigure}
  \hfill
  \begin{subfigure}[t]{0.5\textwidth}
    \includegraphics[height=0.21\textheight]{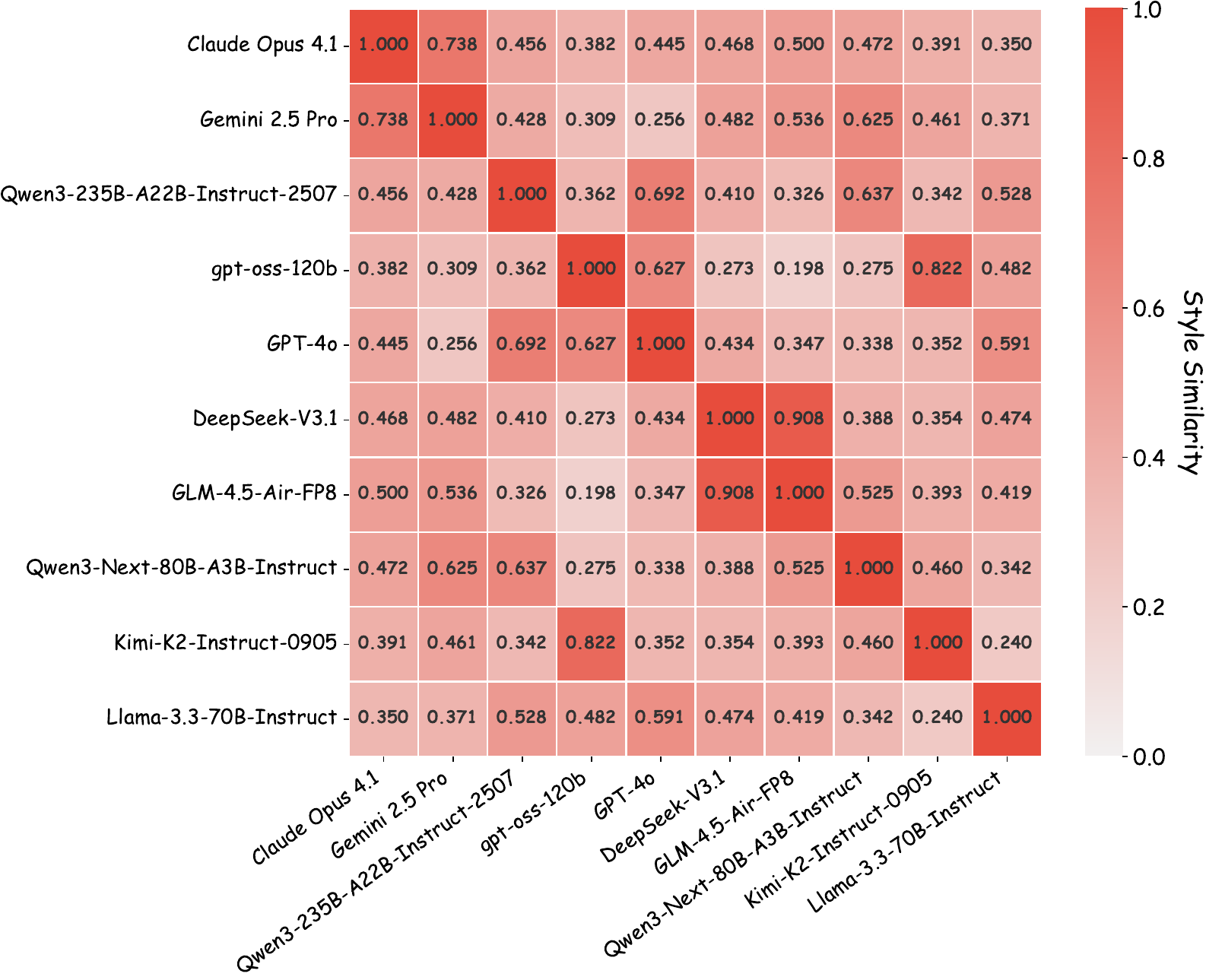}%
    \caption{Stylometric similarity among generators.}
    \label{fig:txt_fake_source_style_similarity}
  \end{subfigure}
  \caption{Data statistics of the text modality in IVT-Set. (a) Category distribution across 7 domains. (b) Generator distribution of fake texts from 10 LLMs. (c) Stylometric similarity among real-text sources. (d) Stylometric similarity among fake-text generators. For (c) and (d), each source is represented by a category-averaged profile of 20 rule-based stylometric features. Pairwise similarity is computed as the Pearson correlation between standardized source profiles and rescaled to $[0,1]$, with higher values indicating more similar profiles.}
  \label{fig:data_stat_text}
\end{figure}

\subsubsection{Detailed Statistics of IVT-Set}
\noindent\textbf{Image.}
The image subset contains 54,054 samples, including 25,234 real images and 28,820 fake images.
It covers 13 broad semantic categories (\cref{fig:img_category_dist}), ranging from common object categories (e.g., animals and vehicle) to more challenging scenarios (e.g., landscape and art).
The category distribution is centered on subjects commonly encountered in real-world images and also includes different scenes and visual styles.
As shown in~\cref{fig:img_gen_dist}, fake images are broadly distributed across different generators, with the largest single source accounting for only 9.5\% of the subset, thereby reducing potential generator-specific bias.
Real and fake images also exhibit closely matched mean quality scores across all categories (\cref{fig:img_quality_score}), with a maximum difference of only 0.18 on a five-point scale.
Meanwhile, both groups contain images spanning a wide range of quality levels.
Overall, the image subset combines comparable quality between real and fake samples with substantial quality diversity within each group,
encouraging detectors to learn forgery-related evidence rather than overall visual quality.

\noindent\textbf{Video.}
The video subset contains 49,470 samples, evenly divided into 24,735 real videos and 24,735 fake videos.
It spans 11 semantic categories (\cref{fig:vid_category_dist}), covering human activities, 
indoor and outdoor scenes, and a variety of subjects commonly found in real-world video content.
The fake videos are generated by 14 models, including both open-source models and commercial platforms, 
and cover a range of generation architectures, strategies, and quality levels (\cref{fig:vid_gen_dist}). 
Each generator contributes at least 806 videos, ensuring sufficient representation across different sources.
We further characterize video quality from two complementary aspects: aesthetic quality and temporal consistency. 
As shown in~\cref{fig:vid_quality_score}, real and fake videos have very similar mean score pairs, with $(0.576, 0.945)$ for real videos and $(0.563, 0.950)$ for fake videos, respectively.
Overall, the video subset combines diverse semantic content and generator sources with closely matched quality between real and fake videos, 
providing a balanced basis for evaluating video forgery detection.

\noindent\textbf{Text.}
The text subset contains 48,672 samples, evenly divided into 24,336 real texts and 24,336 fake texts.
It spans seven writing domains (\cref{fig:txt_category_dist}), covering factual and academic writing, conversational exchanges, 
subjective reviews, creative writing, and news reporting.
Within each domain, real and fake texts are represented by exactly the same number of samples, 
maintaining a consistent domain composition between the two groups.
The fake texts are produced by 10 LLMs, with each model accounting for 6.1\% to 11.2\% of the fake subset (\cref{fig:txt_gen_dist}).
This relatively even distribution provides broad coverage of different LLM output styles without allowing any single model to dominate the subset.
The pairwise stylometric similarity matrices further reveal distinct cross-source patterns in both real and fake texts (\cref{fig:txt_real_source_style_similarity,fig:txt_fake_source_style_similarity}), demonstrating the stylistic diversity of the subset.
Overall, the text subset combines domain-level balance, broad coverage of LLM sources, and source-level stylistic diversity, supporting the evaluation of model-agnostic linguistic evidence across writing domains and generators.

\subsection{Chain-of-Thought Annotation Prompts}
\begin{figure}[!t]
  \centering
  \includegraphics[width=\textwidth,height=\dimexpr\textheight-2.5em\relax,keepaspectratio]{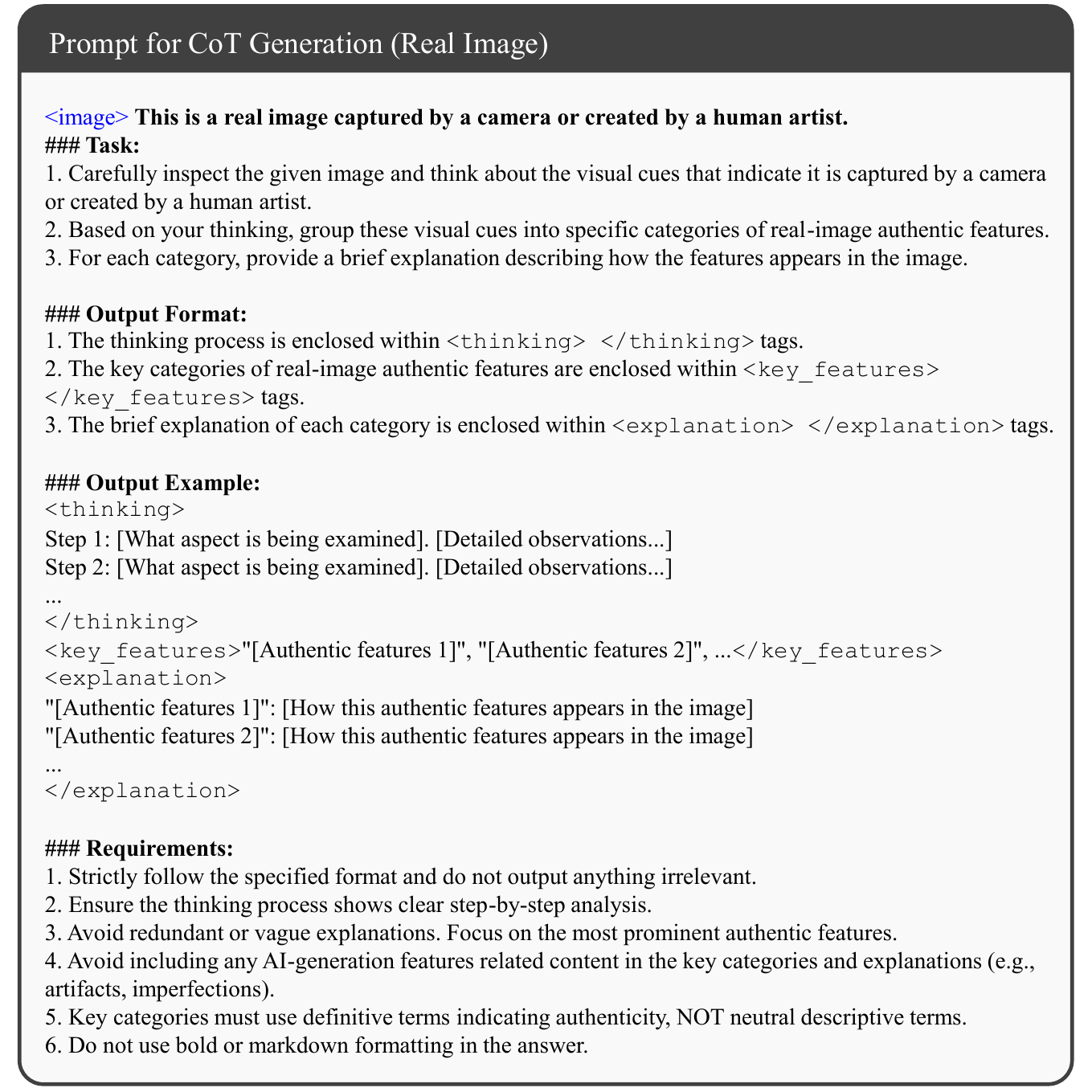}
  \caption{Prompt template for CoT Generation on \textbf{real images}.}
  \label{fig:prompt_cot_real_image}
\end{figure}

\begin{figure}[!t]
  \centering
  \includegraphics[width=\textwidth,height=\dimexpr\textheight-2.5em\relax,keepaspectratio]{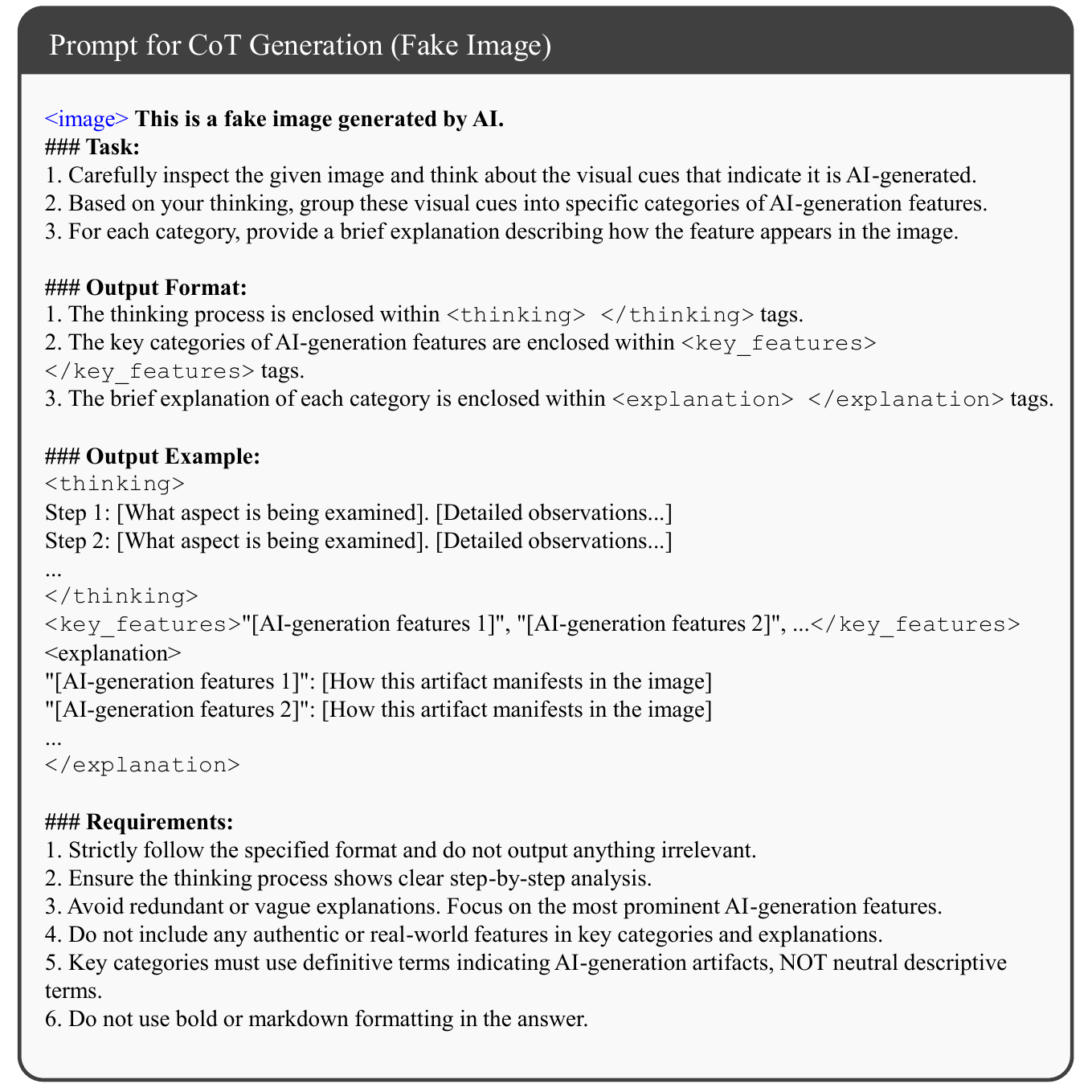}
  \caption{Prompt template for CoT Generation on \textbf{fake images}.}
  \label{fig:prompt_cot_fake_image}
\end{figure}

\begin{figure}[!t]
  \centering
  \includegraphics[width=\textwidth,height=\dimexpr\textheight-2.5em\relax,keepaspectratio]{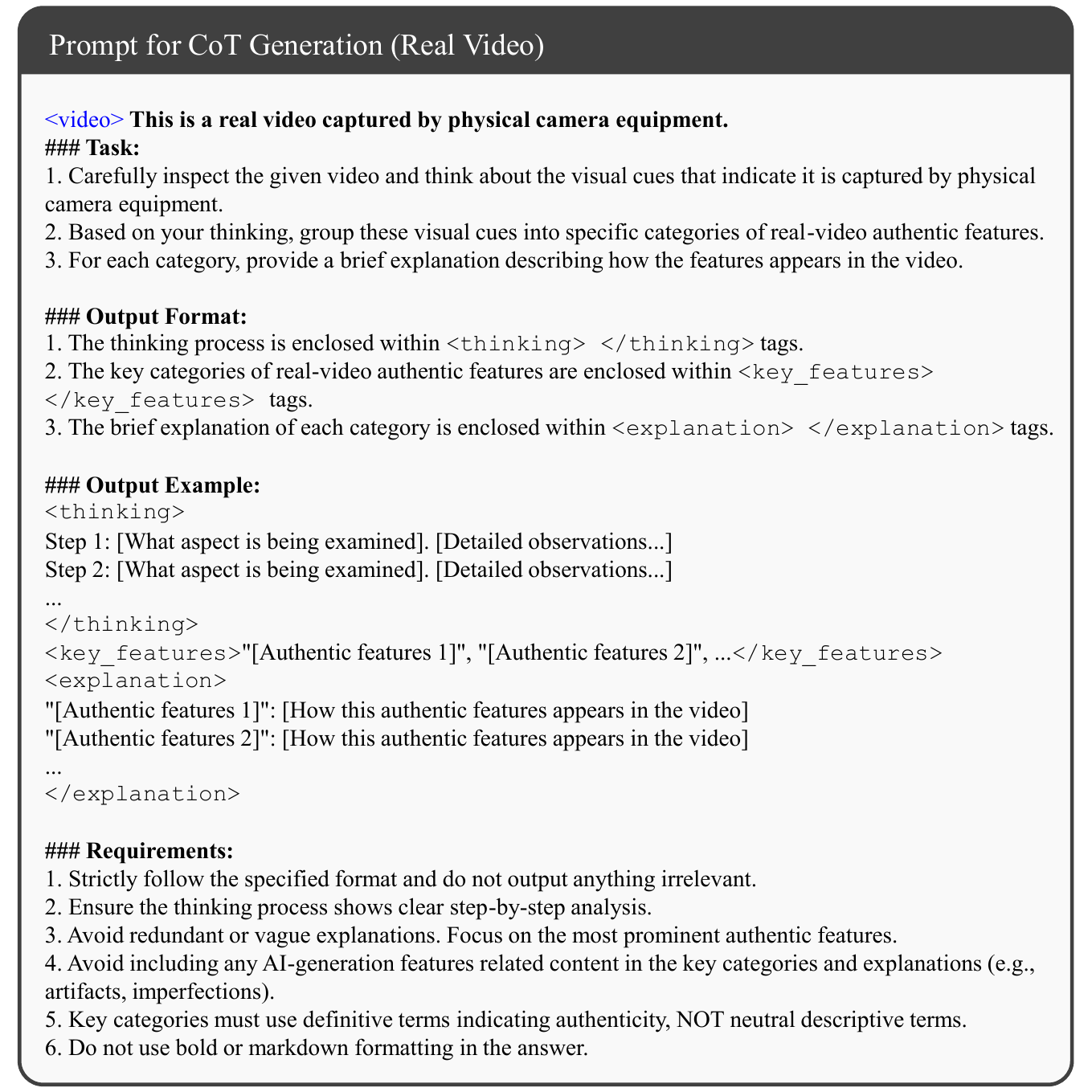}
  \caption{Prompt template for CoT Generation on real videos.}
  \label{fig:prompt_cot_real_video}
\end{figure}

\begin{figure}[!t]
  \centering
  \includegraphics[width=\textwidth,height=\dimexpr\textheight-2.5em\relax,keepaspectratio]{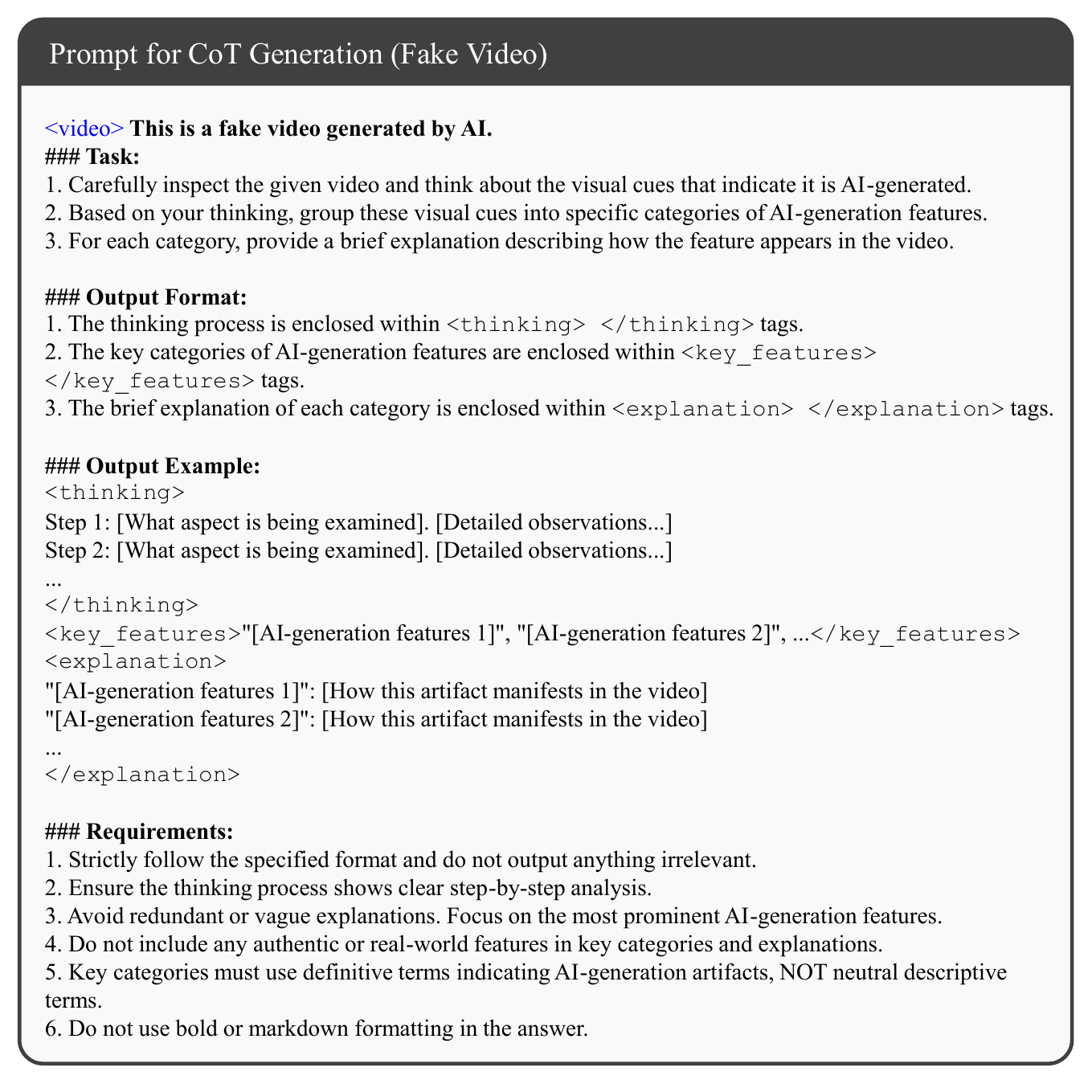}
  \caption{Prompt template for CoT Generation on \textbf{fake videos}.}
  \label{fig:prompt_cot_fake_video}
\end{figure}

\begin{figure}[!t]
  \centering
  \includegraphics[width=\textwidth,height=\dimexpr\textheight-2.5em\relax,keepaspectratio]{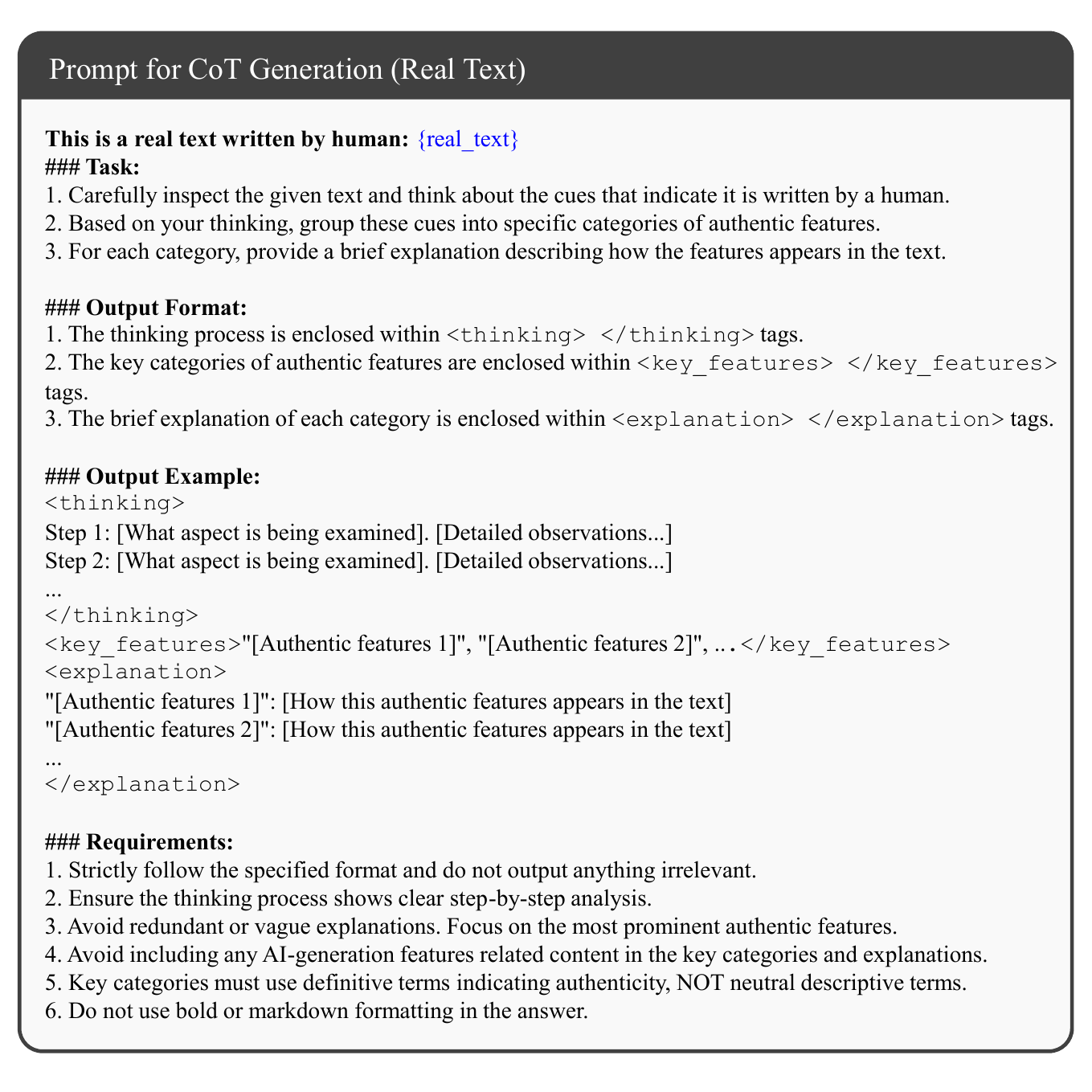}
  \caption{Prompt template for CoT Generation on \textbf{real texts}.}
  \label{fig:prompt_cot_real_text}
\end{figure}

\begin{figure}[!t]
  \centering
  \includegraphics[width=\textwidth,height=\dimexpr\textheight-2.5em\relax,keepaspectratio]{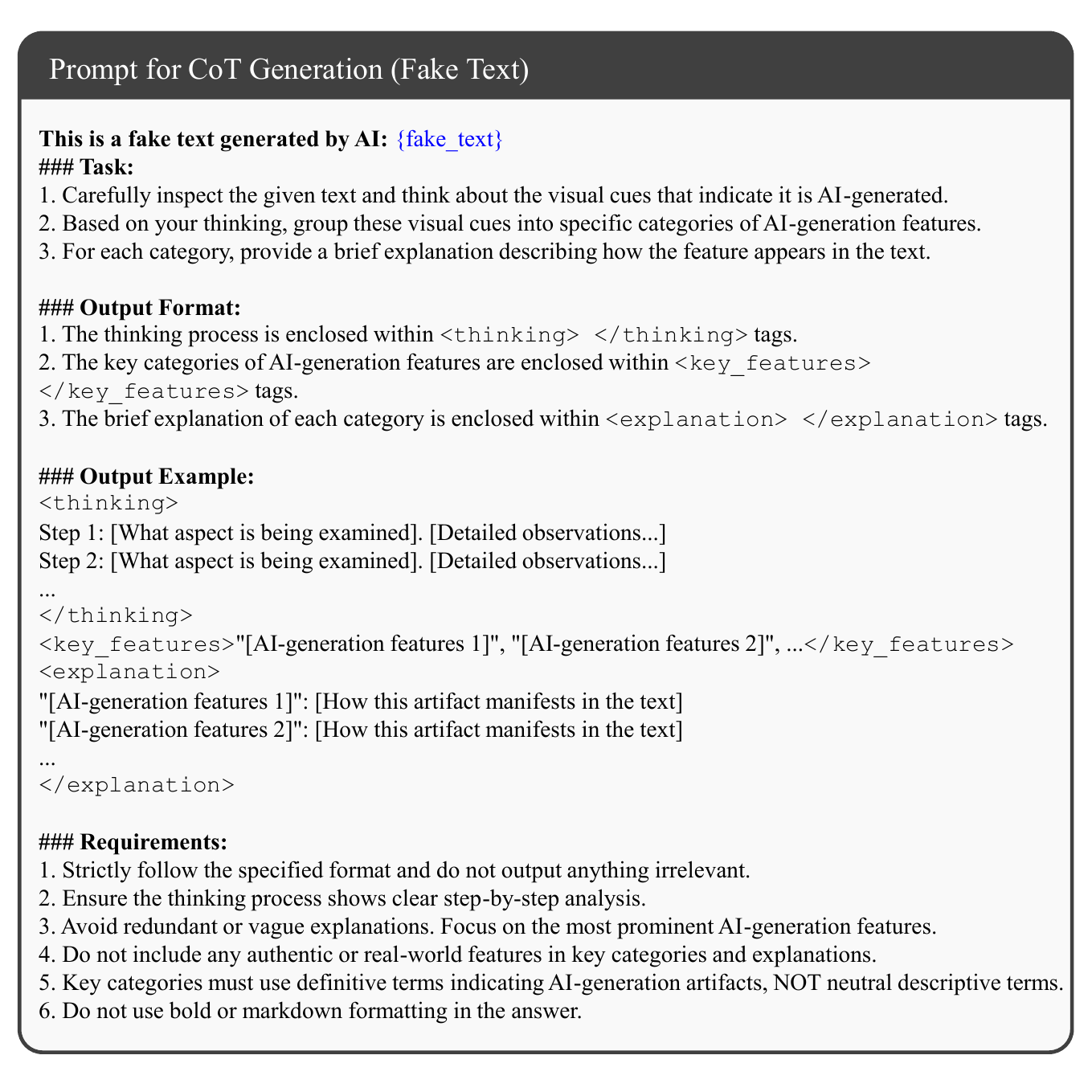}
  \caption{Prompt template for CoT Generation on \textbf{fake texts}.}
  \label{fig:prompt_cot_fake_text}
\end{figure}

\begin{figure}[!t]
\centering
\includegraphics[width=\textwidth,height=\dimexpr\textheight-2.5em\relax,keepaspectratio]{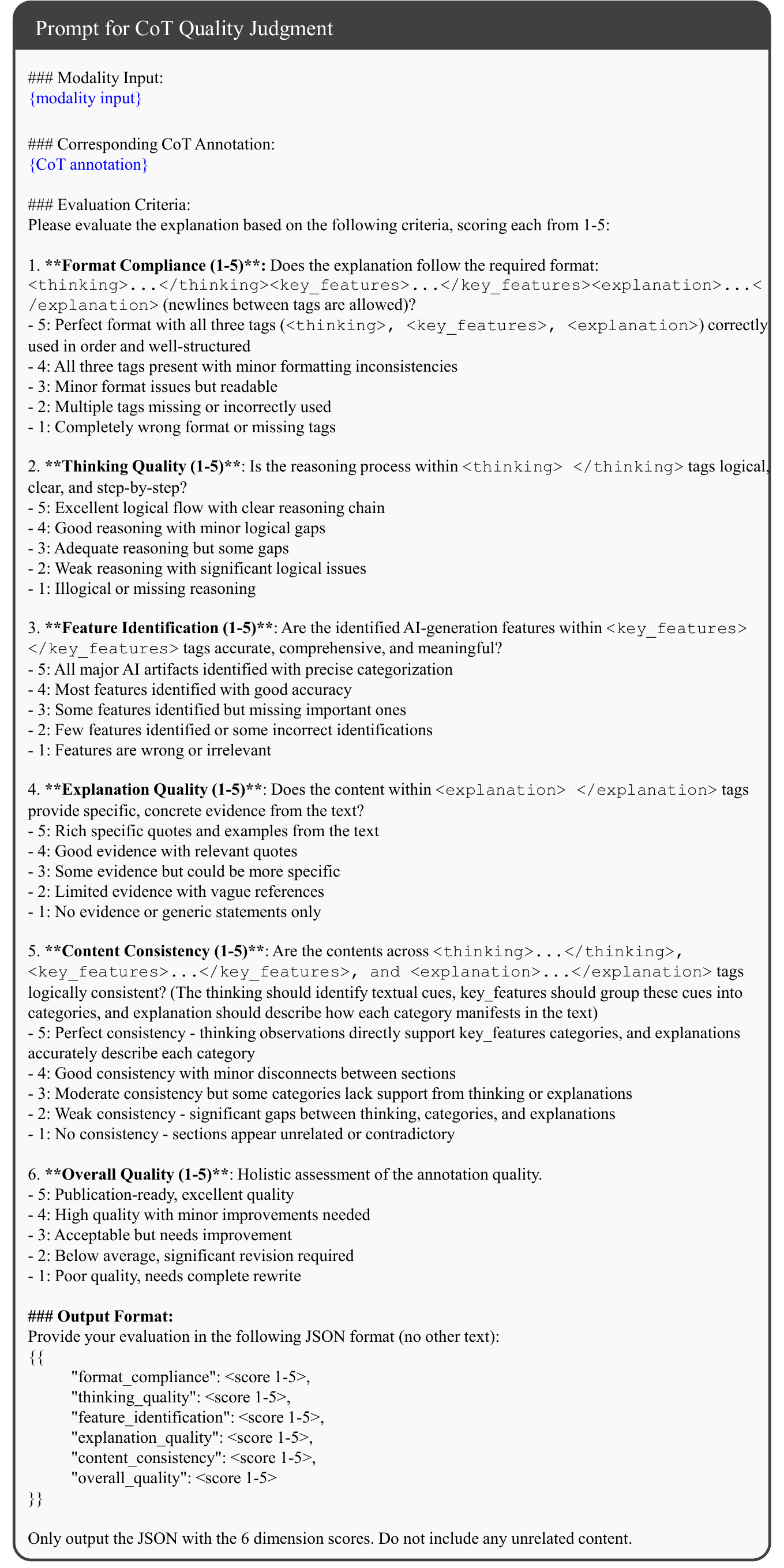}
  \caption{Prompt template for CoT Quality Judgment.}
  \label{fig:prompt_cot_quality}
\end{figure}

\begin{figure}[!t]
\centering
\includegraphics[width=\textwidth,height=\dimexpr\textheight-2.5em\relax,keepaspectratio]{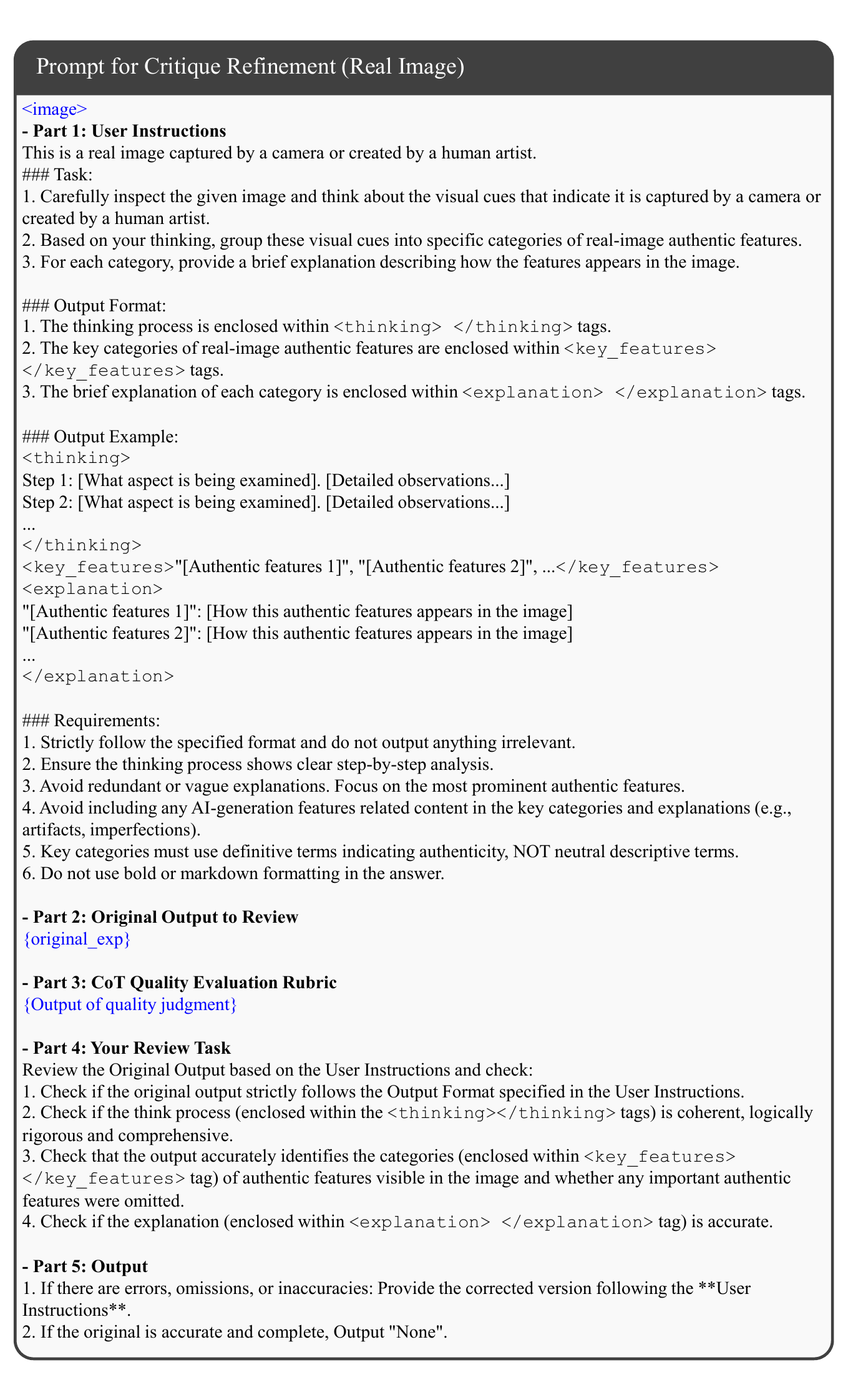}
  \caption{Prompt template for Critique Refinement on \textbf{real images}.}
  \label{fig:prompt_critique_real_image}
\end{figure}

\begin{figure}[!t]
\centering
\includegraphics[width=\textwidth,height=\dimexpr\textheight-2.5em\relax,keepaspectratio]{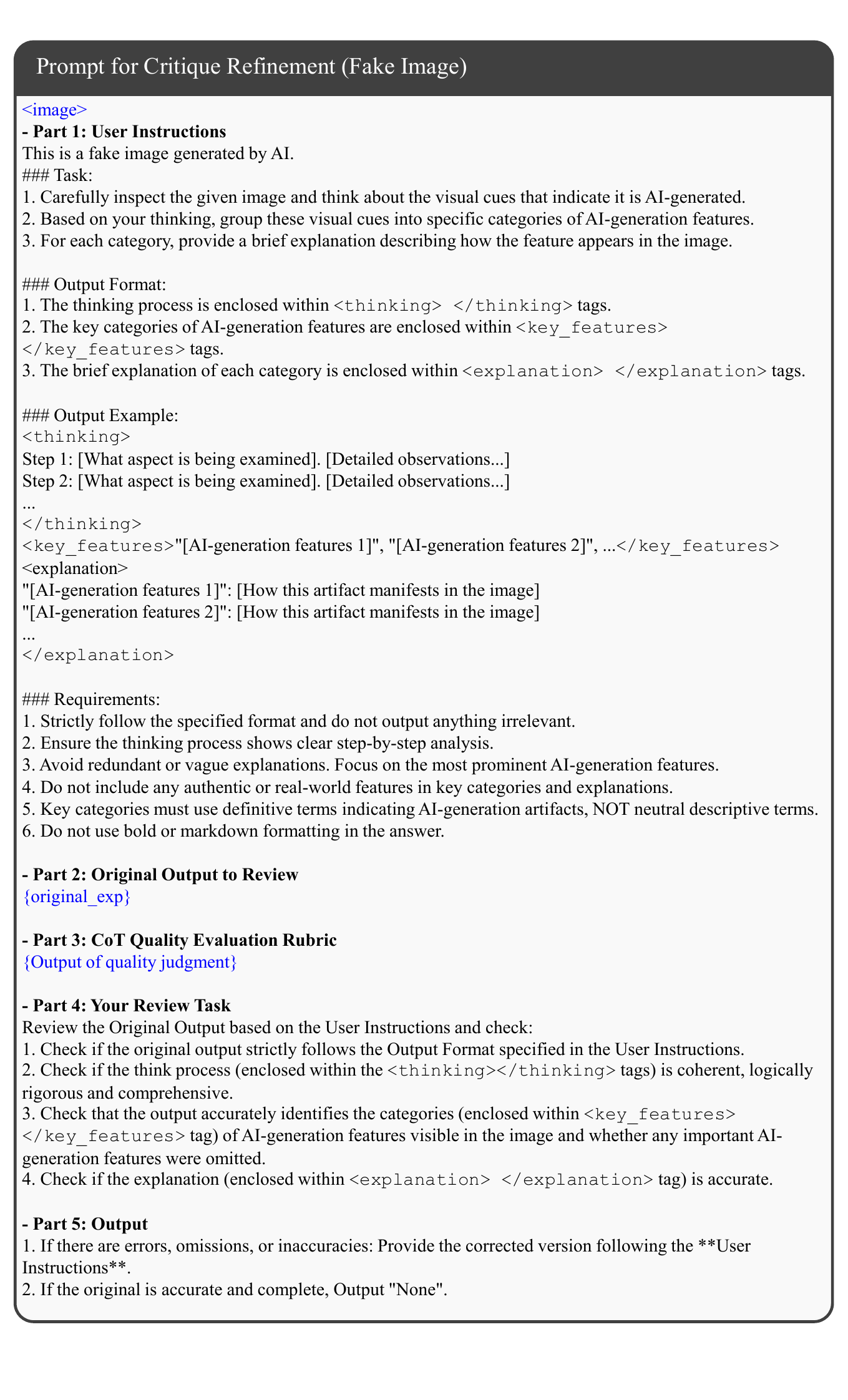}
  \caption{Prompt template for Critique Refinement on \textbf{fake images}.}
  \label{fig:prompt_critique_fake_image}
\end{figure}

\begin{figure}[!t]
\centering
\includegraphics[width=\textwidth,height=\dimexpr\textheight-2.5em\relax,keepaspectratio]{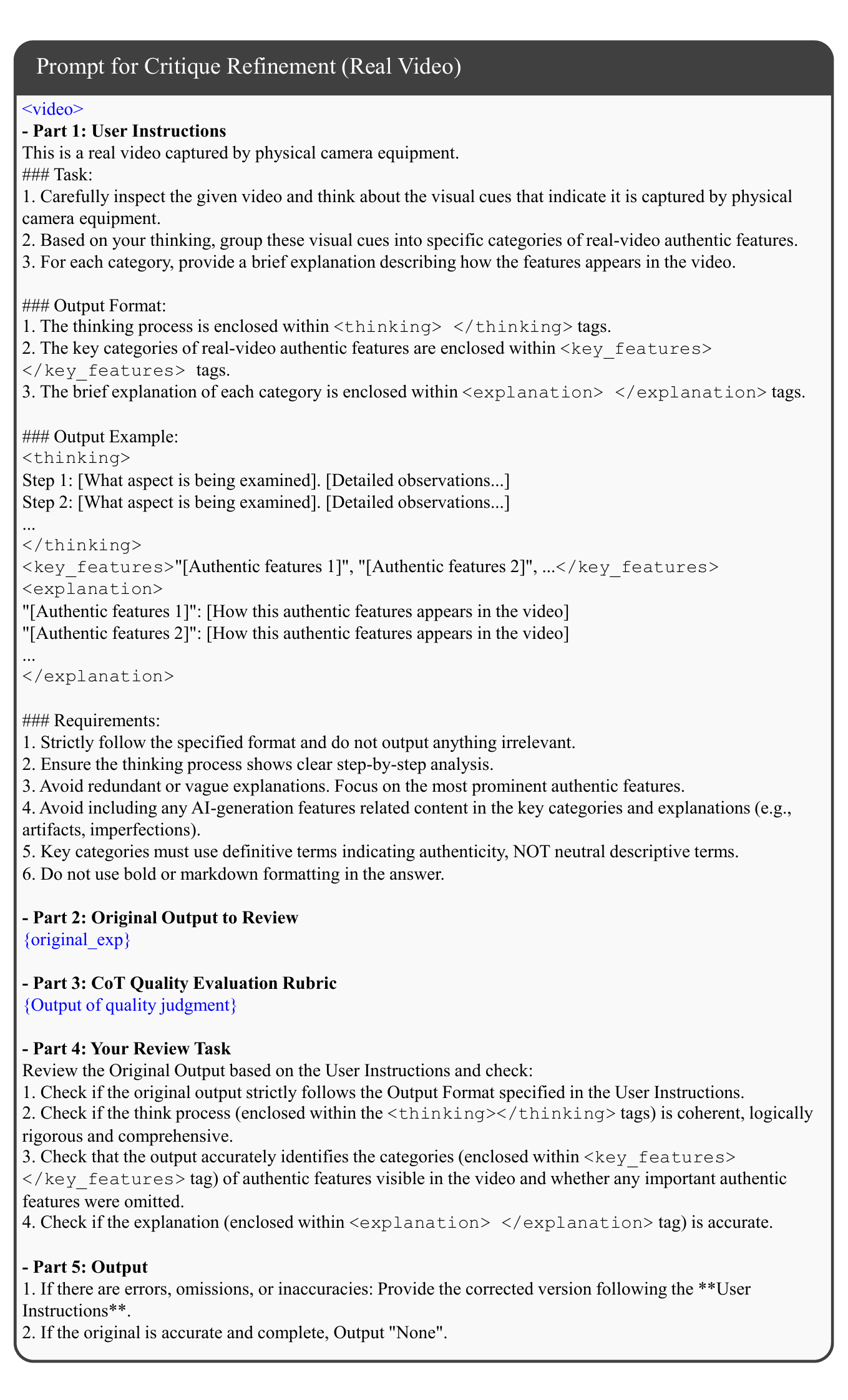}
  \caption{Prompt template for Critique Refinement on \textbf{real videos}.}
  \label{fig:prompt_critique_real_video}
\end{figure}

\begin{figure}[!t]
\centering
\includegraphics[width=\textwidth,height=\dimexpr\textheight-2.5em\relax,keepaspectratio]{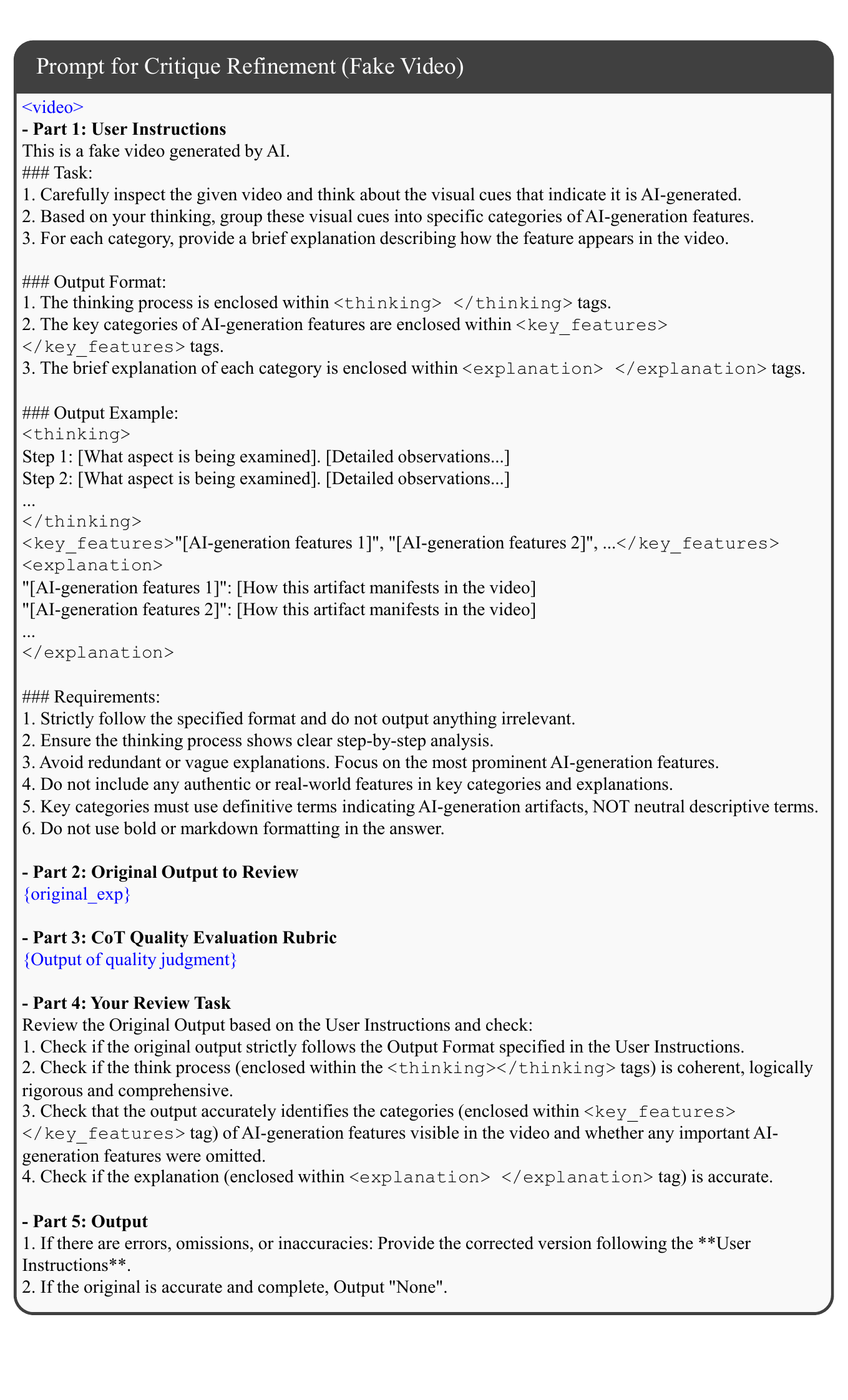}
  \caption{Prompt template for Critique Refinement on \textbf{fake videos}.}
  \label{fig:prompt_critique_fake_video}
\end{figure}

\begin{figure}[!t]
\centering
\includegraphics[width=\textwidth,height=\dimexpr\textheight-2.5em\relax,keepaspectratio]{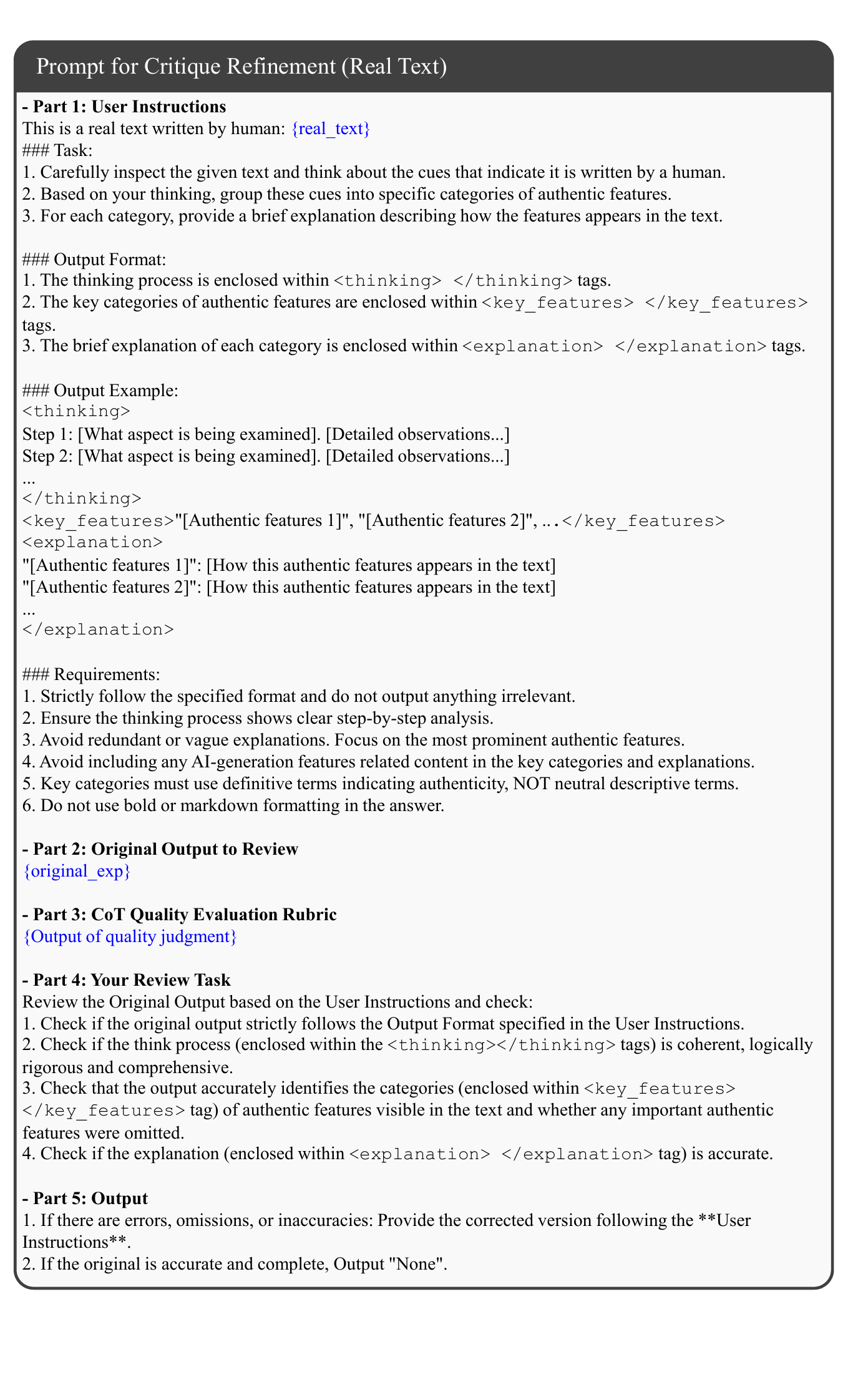}
  \caption{Prompt template for Critique Refinement on \textbf{real texts}.}
  \label{fig:prompt_critique_real_text}
\end{figure}

\begin{figure}[!t]
\centering
\includegraphics[width=\textwidth,height=\dimexpr\textheight-2.5em\relax,keepaspectratio]{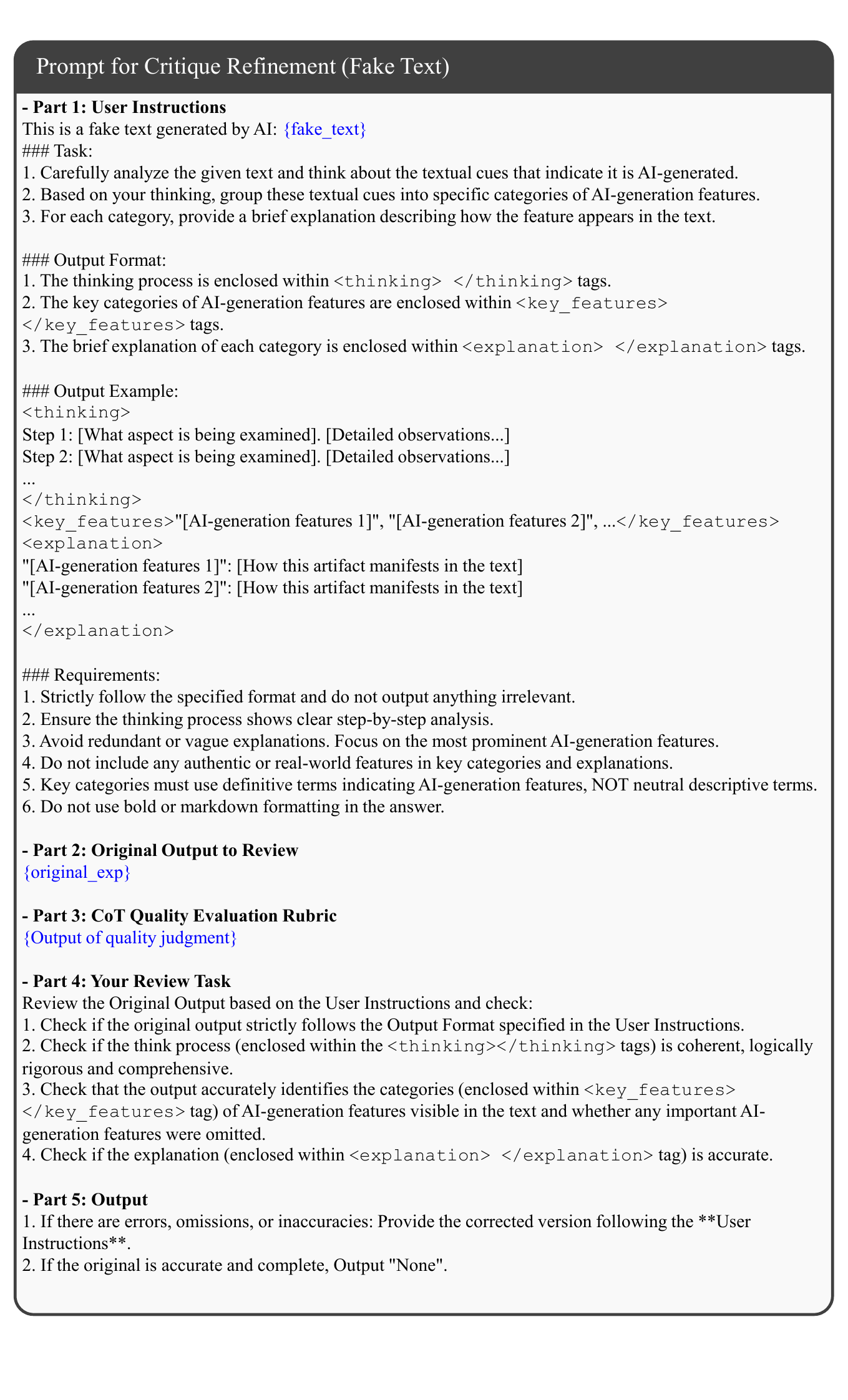}
  \caption{Prompt template for Critique Refinement on \textbf{fake texts}.}
  \label{fig:prompt_critique_fake_text}
\end{figure}

To generate high-quality chain-of-thought (CoT) annotations, we use
modality- and label-specific prompts at different stages.
In the CoT Generation stage, the prompt templates
(\cref{fig:prompt_cot_real_image,fig:prompt_cot_fake_image,fig:prompt_cot_real_video,fig:prompt_cot_fake_video,fig:prompt_cot_real_text,fig:prompt_cot_fake_text})
guide the MLLM to generate an initial CoT annotation according to the
task requirements.
In the CoT Quality Judgment stage, a quality-judgment prompt
(\cref{fig:prompt_cot_quality}) is used with multiple MLLMs or LLMs to
score each annotation along multiple dimensions.
In the Critique--Refinement Loop, dedicated prompt templates
(\cref{fig:prompt_critique_real_image,fig:prompt_critique_fake_image,fig:prompt_critique_real_video,fig:prompt_critique_fake_video,fig:prompt_critique_real_text,fig:prompt_critique_fake_text})
critique and refine the current CoT annotation. The revised annotation
is then re-evaluated by the same MLLM/LLM judges. This loop continues
until the average score reaches the acceptance threshold or three
refinement rounds have been completed.

\end{document}